\documentclass[11pt]{article}

\usepackage[final]{acl}

\usepackage{times}
\usepackage{latexsym}
\usepackage{arydshln}
\usepackage{amsmath}

\usepackage{pgfplots}
\usepackage{xcolor}
\usepackage{enumitem}

\usepackage[T1]{fontenc}
\usepackage{subcaption}
\usepackage[utf8]{inputenc}
\usepackage{microtype}
\usepackage{booktabs}
\usepackage{inconsolata}

\usepackage{graphicx}
\usepackage{tikz}

\usetikzlibrary{shapes.geometric, arrows.meta, positioning, shadows}

\pgfplotsset{compat=1.18}

\definecolor{color_ar}{HTML}{1f77b4}
\definecolor{color_bn}{HTML}{aec7e8}
\definecolor{color_deva}{HTML}{ffbb78}
\definecolor{color_en}{HTML}{98df8a}
\definecolor{color_gu}{HTML}{ff9896}
\definecolor{color_ka}{HTML}{c5b0d5}
\definecolor{color_ml}{HTML}{8c564b}
\definecolor{color_or}{HTML}{e377c2}
\definecolor{color_pa}{HTML}{7f7f7f}
\definecolor{color_sat}{HTML}{bcbd22}
\definecolor{color_ta}{HTML}{17becf}
\definecolor{color_te}{HTML}{9edae5}

\title{MUTANT: A Recipe for Multilingual Tokenizer Design}

\author{Souvik Rana, Arul Menezes, Ashish Kulkarni, Chandra Khatri, Shubham Agarwal \\
        Krutrim  AI, Bangalore, India \\ \{souvik.rana, ashish.kulkarni, shubham.agarwal1\}@olakrutrim.com}

\begin{document}
\maketitle
\begin{abstract}
Tokenizers play a crucial role in determining the performance, training efficiency, and the inference cost of Large Language Models (LLMs). Designing effective tokenizers for multilingual LLMs is particularly challenging due to diverse scripts and rich morphological variation. While subword methods like Byte Pair Encoding (BPE) are widely adopted, their effectiveness in multilingual settings remains underexplored.
We present MUTANT, a recipe for building multilingual tokenizers, with careful vocabulary and training data design, language-aware pre-tokenization, and 
subword and multiword aware training. We also introduce MUTANT-Indic, a tokenizer for India-specific multilingual LLMs, that produces linguistically coherent tokens and achieves state-of-the-art performance.
Evaluated across English, $22$ Indian languages and code data, our tokenizer improves the average fertility score by $39.5\%$ over LLaMA4 and by $18\%$ over Sutra (the current best). This translates to $44\%$ improvement in inference throughput over LLaMA4 while maintaining comparable performance on English and Indic benchmarks. We present detailed ablations across tokenizer training data size, vocabulary size, merging techniques, and pre-tokenization strategies, demonstrating the robustness of our design choices.
\end{abstract}

\section{Introduction}
\label{sec:intro}
Large Language Models (LLMs) \citep{touvron2023llama,grattafiori2024llama, abdin2025phi, guo2025deepseek, yang2025qwen3, team2025gemma} rely on the crucial step of tokenization, the process of converting raw text into discrete units called \textit{tokens}. 
Among the many proposed approaches, subword tokenization schemes such as BPE \citep{sennrich-etal-2016-neural}, Unigram \citep{kudo-2018-subword}, WordPiece \citep{song2021fast}, and their byte-level extensions have become the de facto choice. 
However, tokenization remains an understudied topic 
within the LLM literature~\citep{dagan2024getting, mielke2021between}, 
 especially in multilingual settings~\citep{petrov2023language}, where, skewed fertility scores across languages, often lead to concerns around fairness, high inference latency, cost 
and context size. 
For instance, with the 22 languages listed in the Eighth Schedule of the Constitution of India\footnote{\url{https://en.wikipedia.org/wiki/Languages_with_official_recognition_in_India}}, these issues are especially pronounced for Indic languages comprising multiple scripts and a rich morphology. 
A key metric for evaluating tokenizers is the ``fertility score'' (or token-to-word ratio) \citep{ali2024tokenizer} where, a lower fertility score is desirable due to more efficient (and hence cheaper) LLM training and inference. Our analysis suggests that tokenizers of popular 
multilingual LLMs~\citep{team2025gemma,openai2025gptoss120bgptoss20bmodel,meta2025llama4}, largely designed for English, could produce fertility scores as high as 10.5 (LLaMA-4 tokenizer for Oriya; Table \ref{tab:word-tok-ratio}) for Indic languages, far worse than the near-ideal scores achieved for English. This leads to longer token sequences, higher compute overheads, and poor alignment with linguistic units.

\begin{figure*}[htbp]
    \centering
\includegraphics[width=0.89\linewidth]{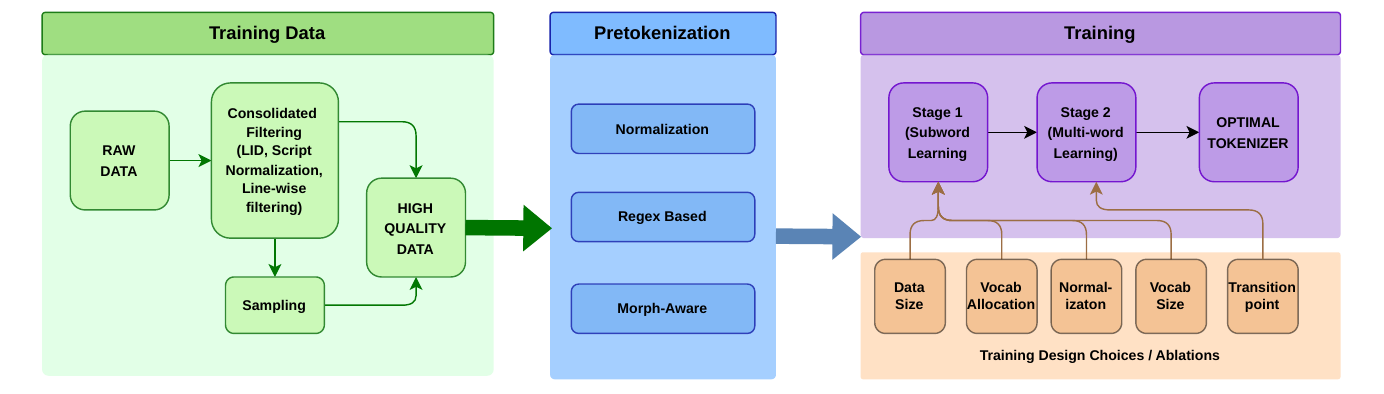}

\caption{MUTANT: Multilingual tokenizer training recipe. Raw text is filtered and consolidated before a two-stage training process consisting of subword and multi-word learning with ablations to identify efficient design choices.}
    \label{fig:mnt-flow}
\end{figure*}


Designing an efficient tokenizer involves making careful choices around the the tokenization algorithm, size of the vocabulary (of tokens) as well as tokenizer training data. 
In this work, we address five core questions and investigate how to design effective multilingual tokenizers by i) examining trade-offs between low- and high-resource languages, ii) joint versus language-specific training, iii) multilingual data balancing, iv) the role of pre-tokenization, and v) the benefits of incorporating multi-word expressions via curriculum training.
Supported by controlled experiments and systematic ablations, we present MUltilingual Tokenizer optimization ANd Training (MUTANT) as a practical and reproducible recipe (see Figure \ref{fig:mnt-flow}) for building equitable, efficient, and linguistically grounded multilingual tokenizers for LLMs. We apply our recipe to train MUTANT-Indic, a state-of-the-art tokenizer for Indic LLMs, demonstrating the benefits of our methodology in a real-world multilingual setting. 
MUTANT-Indic combines linguistically grounded pre-tokenization with a two-stage subword–multi-word learning process~\citep{liu2025superbpespacetravellanguage}, yielding a compact and semantically faithful vocabulary. Figure~\ref{fig:teaser} illustrates some examples where our approach avoids fragmenting common words or idiomatic phrases into unnatural subunits across different languages.
We make the following contributions:

\begin{itemize}[noitemsep]
\item We propose MUTANT, a principled and general recipe for training optimal multilingual tokenizers. We systematically study the effect of vocabulary size, data distribution, language-specific pre-tokenization and multi-stage training on multilingual performance.
\item We demonstrate the effectiveness by training MUTANT-Indic, that achieves state-of-the-art (SOTA) performance for Indic languages.
\item To the best of our knowledge, we are the first to carry out a comprehensive benchmarking of tokenizers across multiple intrinsic and downstream LLM performance measures 
in both pretraining from scratch as well as continual pretraining settings. We will publicly release the evaluation framework to enable reproducibility and community benchmarking. 
\end{itemize}

\section{Related Work}
\label{sec:related-work}

\paragraph{Multilingual Tokenizers.} 
Multilingual tokenization faces challenges from script diversity, morphology, and structural variation. Comparative studies show that vocabulary size and construction strategies strongly affect performance for morphologically rich languages \citep{karthika2025multilingual}, while inefficiencies in underrepresented ones, such as Ukrainian, translate to higher fertility and computational costs \citep{maksymenko2025tokenization}. Tokenization also influences how multilingual models encode morphology, as demonstrated in mT5 vs.\ ByT5 \citep{dang2024tokenization}. For Indic languages, tailored resources \citep{kakwani2020indicnlp} and IndicBERT \citep{ai4bharat2022indicbert} highlight the value of domain-specific tokenization. Recent benchmarks further reveal economic implications, with BLOOM's tokenizer achieving the best cost efficiency among popular multilingual LLMs \citep{adasci2024multilingual}. Together, these studies show that current multilingual tokenizers fragment low-resource and morphologically rich languages, motivating approaches that combine normalization, language-tailored pre-tokenization, and multi-word learning to achieve better efficiency and fairness.


\begin{figure*}[htbp]
\centering
\begin{tikzpicture}
\begin{axis}[
    xbar stacked,
    width=\textwidth,
    height=2.5cm, 
    bar width=0.7cm,
    xmin=0, xmax=200000,
    xlabel={Vocabulary Size},
    ytick=\empty, 
    y axis line style={opacity=0}, 
    axis on top,
    enlarge y limits=0.5,
    xtick pos=bottom,
    scaled x ticks=false,
    xtick={0, 25000, 50000, 75000, 100000, 125000, 150000, 175000, 200000},
    xticklabel style={
        /pgf/number format/fixed,
        /pgf/number format/1000 sep={} 
    },
    legend style={
        at={(0.5,-0.7)}, 
        anchor=north,
        legend columns=12, 
        draw=black!20, 
        font=\scriptsize,
        /tikz/every even column/.append style={column sep=0.15cm}
    }
]

\addplot[fill=color_ar, draw=none] coordinates {(4000,0)};
\addplot[fill=color_bn, draw=none] coordinates {(10000,0)};
\addplot[fill=color_deva, draw=none] coordinates {(48000,0)};
\addplot[fill=color_en, draw=none] coordinates {(70000,0)};
\addplot[fill=color_gu, draw=none] coordinates {(6000,0)};
\addplot[fill=color_ka, draw=none] coordinates {(12000,0)};
\addplot[fill=color_ml, draw=none] coordinates {(12000,0)};
\addplot[fill=color_or, draw=none] coordinates {(6000,0)};
\addplot[fill=color_pa, draw=none] coordinates {(6000,0)};
\addplot[fill=color_sat, draw=none] coordinates {(2000,0)};
\addplot[fill=color_ta, draw=none] coordinates {(12000,0)};
\addplot[fill=color_te, draw=none] coordinates {(12000,0)};

\legend{ar, bn, deva, en, gu, ka, ml, or, pa, sat, ta, te}

\end{axis}
\end{tikzpicture}
\caption{Vocabulary size distribution across language scripts. See Appendix \ref{app:iso} for script details.}
\label{fig:script-vocab-size}
\end{figure*}

\paragraph{Tokenization Algorithms.} 

Tokenization strategies differ in both theory and practice. While alternate sub-word tokenization algorithms have been explored in the past such as WordPiece \citep{song2021fast}, Unigram LM \citep{kudo2018sentencepiece}, Byte Pair Encoding (BPE) remains the most widely adopted. Originally developed for compression \citep{gage1994new} and later adapted for neural MT \citep{sennrich2016neural}, BPE merges frequent character pairs to balance coverage with efficiency. Its variants aim to address inefficiencies: PickyBPE \citep{chizhov2024pickybpe} discards uninformative merges to improve vocabulary utility, while Scaffold-BPE \citep{lian2024scaffold} iteratively prunes low-frequency scaffold tokens to reduce imbalance and enhance downstream performance. Recent extensions like SuperBPE \citep{liu2025superbpe} expand beyond word boundaries, jointly learning subwords and multi-words yielding improved compression and inference efficiency in a 2-stage curriculum. BoundlessBPE~\citep{schmidt2024boundless}, another contemporary work, relaxes the Pre-tokenization word boundary constraint in an single stage learning step. Our work compares these two recent approaches and shows that two-stage curriculum
preserves subword coverage while capturing larger semantic units in morphologically rich Indian languages.

\paragraph{Pre-tokenization.} Pre-tokenization plays a pivotal role in shaping token boundaries, directly influencing both compression efficiency and reasoning performance \citep{xue2024getting}. SentencePiece \citep{kudo2018sentencepiece} introduced a language-agnostic approach by treating input as raw streams, effective for languages without whitespace boundaries. More recent approaches like BoundlessBPE \citep{schmidt2024boundless} relax pre-token constraints to improve frequency distributions, while regex-based designs continue to prove crucial for capturing script-specific structures.

\section{MUTANT: Tokenizer Training Recipe}
\label{sec:algo}

Language modeling involves estimating the probability distribution over text sequences, $P(S)$, where $S$ may represent a sentence, paragraph, or document. To achieve this, the text is first converted into a sequence of discrete tokens through a tokenization function $g(S) = X = (X_1, X_2, \ldots, X_n) \in V^n$, where $V$ denotes the vocabulary and $n$ the sequence length. Tokenizers can be open-vocabulary, ensuring any string can be represented (e.g., byte-level), or closed-vocabulary, where unseen text maps to an out-of-vocabulary symbol (e.g., word lists) \citep{rae2021scaling}. In our work, we adopt an open-vocabulary approach that combines byte-pair encoding (BPE) with a UTF-8 byte fallback, following 
~\citet{radford2018improving}.
In this section, we present a systematic recipe for designing multilingual tokenizers that jointly addresses data curation, pre-tokenization, and subword–multiword learning. Figure~\ref{fig:mnt-flow} provides an end-to-end overview of the pipeline, illustrating how raw multilingual text is filtered and sampled, transformed through a fixed pre-tokenization stage, and used to train a two-stage tokenizer with explicit design choices on vocabulary size and allocation, detailed below. 



\subsection{Training Data and Vocabulary}
The composition of the training data and the vocabulary size are design choices that directly impact the quality of tokens
and the tokenizer efficiency.

\textbf{Training data volume:} 
Effective tokenizer training relies on a sufficiently large  carefully curated multilingual corpus; beyond moderate scale, further data scaling yields diminishing returns \citep{reddy2025much} in quality (Section \ref{sec:ablation}).



\begin{table*}[htbp]
\centering
\small

\resizebox{\textwidth}{!}{%
\begin{tabular}{l*{24}{r}}
\toprule
 & as & bn & brx & code & doi & eng & gom & gu & hi & kas & kn & mai & ml & mni & mr & nep & or & pa & san & sat & snd & ta & te & urd \\
\midrule
Size (MB)       & 56 & 562 & 13 & 4 & 1 & 148 & 67 & 83 & 422 & 30 & 252 & 60 & 311 & 34 & 152 & 82 & 46 & 144 & 110 & 0.47 & 38 & 502 & 486 & 38 \\
\# Lines (K)     & 65 & 681 & 19 & 118 & 2 & 449 & 135 & 91 & 545 & 21 & 273 & 129 & 337 & 64 & 257 & 126 & 59 & 198 & 139 & 1 & 69 & 658 & 773 & 27 \\
Avg W/Line      & 51 & 47 & 41 & 3 & 43 & 56 & 37 & 59 & 59 & 145 & 41 & 39 & 35 & 43 & 33 & 40 & 45 & 57 & 36 & 26 & 70 & 32 & 32 & 185 \\
\bottomrule
\end{tabular}%
}
\caption{Evaluation corpus statistics across 22 Indic languages, English, and code in Section \ref{subsec:eval-framework} including size (in MB), number of lines (in K) and average words per line. We report standard ISO codes here (Appendix Section \ref{app:iso}).}
\label{tab:eval-stats}
\end{table*}

\textbf{Data quality and diversity:} 
We combine
diverse, high-quality sources that is representative of natural language use~\citep{hayase2024data}, such as, the Web, curated multilingual datasets, 
Wikipedia, math, code and others. The raw data is further subjected to filtering for language/script mixing, normalization, short documents (or lines) 
and controlled source-wise sampling (see Appendix \ref{appendix:tokenizer-imp} for MUTANT-Indic). 


\textbf{Vocabulary size and multilingual allocation:} In multilingual settings, vocabulary allocation must be explicitly language (or script)-aware to preserve subword and multi-word granularity across both high- and low-resource languages. We determine script-wise vocabulary budgets through ablations that vary allocation across scripts, identifying distributions that balance fragmentation and coverage (Figure~\ref{fig:script-vocab-size}). To realize these target allocations in practice, we evaluate two strategies:
\textit{i) explicit merging}, where we train script-specific tokenizers and concatenates their vocabularies via rule stacking; and
\textit{ii) corpus-driven alignment}, where we train a single tokenizer on a multilingual corpus whose data proportions are aligned with the desired script-wise vocabulary budgets. 

Explicit merging introduces distributional interference and inconsistent segmentation across scripts (Table~\ref{tab:individual-vs-merged-selected}). In contrast, corpus-driven alignment allows the learned vocabulary to naturally converge toward the target script-wise proportions, better mirroring corpus composition (Table~\ref{tab:unified-hypothesis}) and achieving consistently lower fertility across scripts (Table~\ref{tab:word-tok-ratio}), outperforming both explicit merging and public baselines.

\subsection{Tokenizer Training}


We adopt a two-stage curriculum following the framework of \citet{liu2025superbpe}. This procedure forms the training strategy used in MUTANT and its Indic instantiation, MUTANT-Indic.

\begin{table*}[htbp]
\centering
\resizebox{\textwidth}{!}{%
\begin{tabular}{lrrrrrrrrrrrrrrrrrrrrrrrr}
\toprule
 Regex   & as & bn & brx & code & doi & eng & gom & gu & hi & kas & kn & mai & ml & mni & mr & nep & or & pa & san & sat & snd & ta & te & urd \\
\midrule
GPT-2 & 4.36 & 4.72 & 4.67 & 1.57 & 2.88 & 1.32 & 3.95 & 4.12 & 3.47 & 2.47 & 5.95 & 3.17 & 7.08 & 3.30 & 4.86 & 4.37 & 4.44 & 3.28 & 5.97 & 2.71 & 1.30 & 6.53 & 5.61 & 1.29 \\
LLaMA-4 & 1.83 & 1.74 & 1.99 & 1.54 & 1.56 & 1.33 & 2.17 & 1.83 & 1.36 & 1.36 & 2.15 & 1.56 & 2.24 & 2.27 & 1.61 & 1.59 & 1.65 & 1.47 & 2.51 & 3.60 & 1.45 & 2.07 & 1.83 & 1.47 \\
\bottomrule
\end{tabular}%
}
\caption{Fertility scores ($\downarrow$) showing LLaMA-4 regex outperforms GPT-2 in Stage-1 tokenizer training.}
\label{tab:regex}
\end{table*}

\begin{table*}[htbp]
\centering
\resizebox{\textwidth}{!}{
\begin{tabular}{l*{24}{r}}
\toprule
Tokenizer ($\downarrow$) & as & bn & brx & code & doi & eng & gom & gu & hi & kas & kn & mai & ml & mni & mr & nep & or & pa & san & sat & snd & ta & te & urd \\
\midrule
Gemma-3 & 2.65 & \textbf{1.69} & 2.84 & 1.79 & 1.69 & 1.39 & 2.60 & 2.50 & 1.47 & 1.48 & 3.34 & 1.91 & 3.45 & 2.07 & 2.03 & 2.03 & 4.42 & 2.83 & 3.37 & 5.16 & 2.03 & 2.50 & 2.94 & 1.44 \\
GPT-OSS & 2.66 & 2.41 & 3.17 & 1.51 & 1.89 & 1.33 & 2.73 & 2.37 & 1.72 & 1.58 & 3.34 & 2.01 & 3.51 & 2.41 & 2.61 & 2.10 & 6.26 & 2.71 & 3.89 & 13.01 & 1.76 &3.18 & 3.13 & 1.51 \\
LLaMA-4 & 4.40 & 2.93 & 3.34 & 1.46 & 2.00 & 1.34 & 2.84 & 3.37 & 1.83 & 1.72 & 4.23 & 2.28 & 4.95 & 2.73 & 2.79 & 2.46 & 10.51 & 3.23 & 4.12 & 9.04 & 2.13 & 5.87 & 4.53 & 1.76 \\
Sarvam & 4.24 & 1.91 & 2.92 & 2.14 & 1.85 & 1.66 & 3.01 & 2.11 & 1.53 & 1.91 & 2.53 & 2.11 & 3.19 & 4.60 & 1.94 & 2.35 & 2.43 & 1.67 & 3.78 & 13.07 & 7.62 & 2.49 & 2.63 & 7.93 \\
Sutra & 2.12 & 2.07 & 3.06 & 2.12 & 1.78 & 1.17 & 2.68 & 2.15 & 1.62 & 1.48 & 2.71 & 2.08 & 3.10 & 2.40 & 2.18 & 2.01 & 2.24 & 1.50 & 3.76 & \textbf{2.03} & 2.23 & 2.58 & 2.77 & 1.55 \\
\hdashline
MUTANT-Indic & \textbf{1.85} & 1.74 & \textbf{2.04} & 1.47 & \textbf{1.45} & \textbf{1.12} & \textbf{2.17} & \textbf{1.77} & \textbf{1.23} & \textbf{1.21} & \textbf{2.19} & \textbf{1.58} & \textbf{2.30} & 2.28 & \textbf{1.63} & \textbf{1.62} & \textbf{1.65} & \textbf{1.39} & \textbf{2.59} & 3.72 & \textbf{1.45} & \textbf{2.12} & \textbf{1.88} & \textbf{1.44} \\
\bottomrule
\end{tabular}
}
\caption{Fertility score ($\downarrow$)  comparison for Indic focused and good Indic support tokenizers across languages. MUTANT-Indic performs best in 20 of 24 languages. An extended version in Table \ref{tab:big_fertility_scores} (Appendix).}
\label{tab:word-tok-ratio}
\end{table*}

\begin{table*}[htbp]
\centering
\small
\resizebox{\textwidth}{!}{%
\begin{tabular}{l*{24}{r}}
\toprule
Tokenizer ($\uparrow$) & as & bn & brx & code & doi & eng & gom & gu & hi & kas & kn & mai & ml & mni & mr & nep & or & pa & san & sat & snd & ta & te & urd \\
\midrule
Gemma-3     & 6.37 & \textbf{10.45} & 5.87 & 2.33 & 6.75 & 4.36 & 5.29 & 6.31 & 9.16 & 7.01 & 6.73 & 6.42 & 7.57 & 6.23 & 8.90 & 8.31 & 3.76 & 4.62 & 6.66 & 2.59 & 3.87 & 9.60 & 6.82 & 5.55 \\
GPT-oss     & 6.36 & 7.35 & 5.27 & 2.77 & 6.04 & 4.55 & 5.02 & 6.68 & 7.83 & 6.54 & 6.74 & 6.11 & 7.43 & 5.34 & 6.94 & 8.04 & 2.65 & 4.83 & 5.79 & 1.03 & 4.46 & 7.56 & 6.41 & 5.28 \\
LLaMA-4     & 3.84 & 6.05 & 4.99 & \textbf{2.85} & 5.70 & 4.53 & 4.84 & 4.69 & 7.37 & 6.03 & 5.33 & 5.39 & 5.26 & 4.71 & 6.49 & 6.84 & 1.58 & 4.05 & 5.45 & 1.48 & 3.69 & 4.10 & 4.43 & 4.54 \\

Sarvam-2B   & 3.92 & 9.42 & 5.70 & 1.95 & 6.16 & 3.65 & 4.55 & 7.62 & 8.92 & 5.29 & 9.07 & 5.83 & 8.63 & 2.81 & 9.46 & 7.20 & 7.17 & 7.95 & 6.03 & 1.02 & 1.02 & 9.74 & 8.46 & 1.00 \\
Sutra       & 8.04 & 8.50 & 5.44 & 1.97 & 6.39 & 5.15 & 4.98 & 7.36 & 8.33 & 7.00 & 8.38 & 5.88 & 8.75 & 5.36 & 8.35 & 8.45 & 7.73 & 8.76 & 6.04 & \textbf{6.59} & 3.49 & 9.38 & 8.04 & 5.15 \\
\hdashline
MUTANT-Indic & \textbf{9.12} & 10.15 & \textbf{8.18} & 2.84 & \textbf{7.86} & \textbf{5.44} & \textbf{6.29} & \textbf{8.95} & \textbf{11.01} & \textbf{8.59} & \textbf{10.30} & \textbf{7.80} & \textbf{11.33} & \textbf{5.67} & \textbf{11.11} & \textbf{10.39} & \textbf{10.07} & \textbf{9.40} & \textbf{8.70} & 3.60 & \textbf{5.40} & \textbf{11.32} & \textbf{10.70} & \textbf{5.55} \\
\bottomrule
\end{tabular}%
}
\caption{Bytes-per-token score ($\uparrow$)  comparison for Indic focused and good support tokenizers across languages here. MUTANT-Indic performs best in 22 of 24 languages.}
\label{tab:bytes-per-token-transposed}
\end{table*}

\textbf{Stage 1 (Subword Learning):}\label{sec:stage-1-algo} Training begins with standard byte-pair encoding (BPE) applied after whitespace pre-tokenization. 
This ensures that merges occur only within word boundaries, allowing the tokenizer to learn fine-grained \textit{subword units} such as roots, affixes, and common morphemes. Stage 1 continues until the vocabulary reaches a pre-defined \textit{transition point} $t$ ($<|V|$).

\begin{figure}[htbp]
    \centering
    \includegraphics[width=\linewidth]{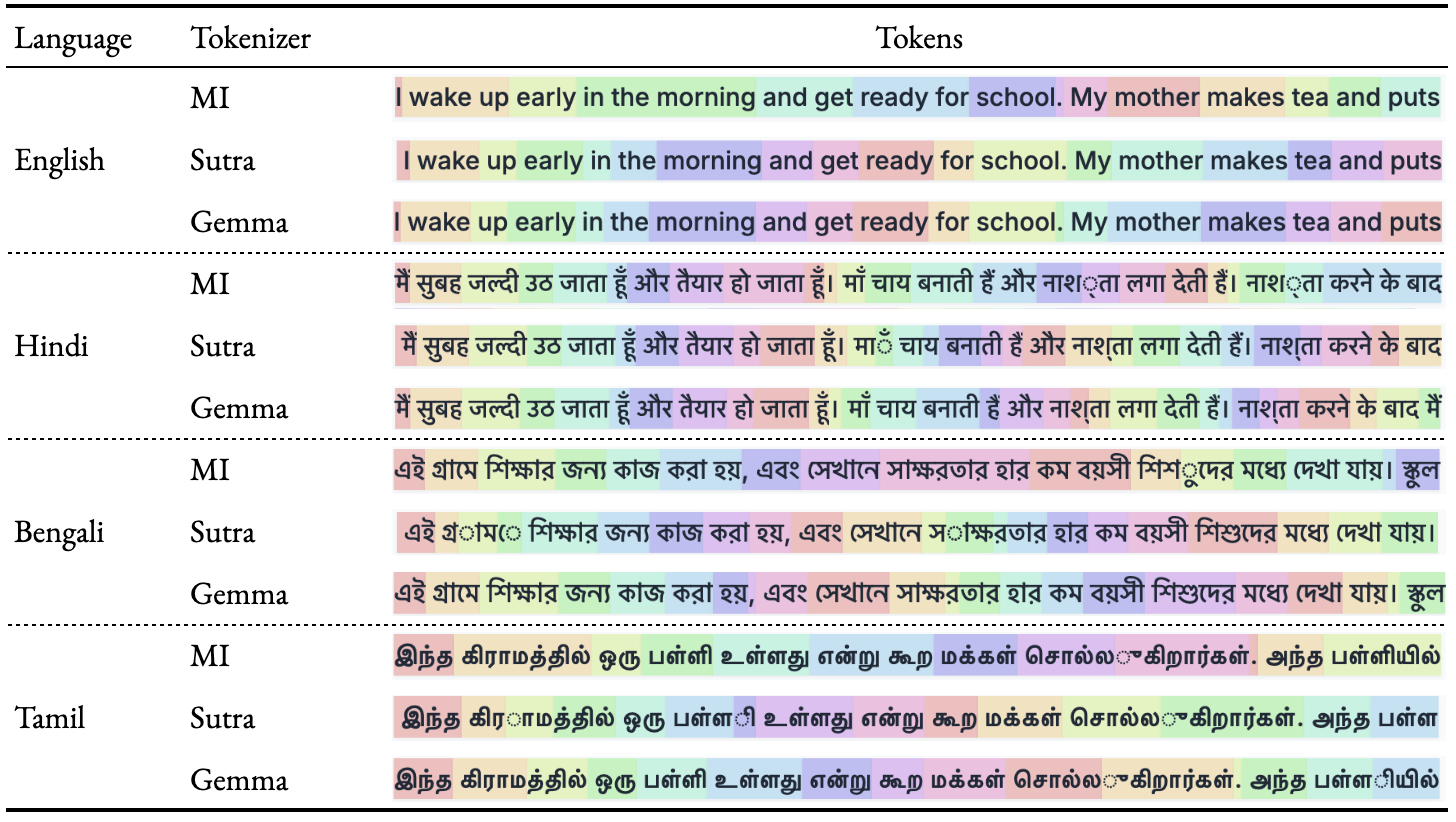}
    \caption{MUTANT-Indic (MI) captures multi-words (e.g. ``wake up", ``in the morning") and avoids fragmenting Indic words (see for e.g. Bengali, Tamil).}
    \label{fig:teaser}
\end{figure}

\textbf{Stage 2 (Multi-word Learning):} After reaching $t$, training resumes without whitespace constraints, allowing BPE to merge across word boundaries. This enables the formation of \textit{multi-words}, frequent multiword expressions or collocations (e.g., ``in the morning'' in Figure \ref{fig:teaser})
improving compression and reducing token counts for common phrases.

This two-stage tokenizer training is particularly effective for 
scripts with complex variations where, meaningful subwords are first anchored and then composed into frequent multiword units. 
\subsection{Pre-tokenization}
As illustrated in Figure~\ref{fig:mnt-flow}, pre-tokenization is a fixed, upstream component of our tokenizer design. 
It segments raw text before subword learning to improve token consistency and efficiency. 
For MUTANT-Indic, we explore
regex-based, Unicode normalization, and morphology-aware strategies. Unicode-aware regex separates punctuation, handles numeric groups, and aligns tokens with semantic units, while NFKC normalization standardizes visually identical characters to reduce orthographic sparsity (Table~\ref{table:norm}). To improve robustness across Indic scripts, we explore GPT-2 pre-tokenization rules against LLaMA-4 regex in Stage 1, where the latter yields
substantially lower token fragmentation and improving token-to-word ratios by 38–40\% on Indic languages (Table~\ref{tab:regex}). 


Script-agnostic segmentation (as in LLaMA-4 regex rules) across Indic as well as non-Indic scripts helps reduce token fragmentation, while 
relaxing whitespace constraints in Stage 2 enables the learning of frequent multi-word expressions (see Appendix~\ref{sec:regex_exp}). However, unconstrained merging can destabilize generation by creating sentence-crossing tokens. Enforcing additional sentence-level boundary constraints in Stage 2 regex prevents this while preserving multi-word modeling benefits.

Additionally, morphology-aware segmentation decomposes words into roots and affixes to capture recurring morphemes. Although we explore morphology-aware segmentation, integrating it into the tokenization pipeline without incurring latency overhead is challenging (Appendix \ref{sec:morph}).

\begin{table*}[htbp]
\centering
\small

\resizebox{\textwidth}{!}{
\begin{tabular}{l*{24}{r}}
\toprule
Tokenizer ($\downarrow$) & as & bn & brx & code & doi & eng & gom & gu & hi & kas & kn & mai & ml & mni & mr & nep & or & pa & san & sat & snd & ta & te & urd \\
\midrule
Gemma-3             & 0.63 & \textbf{0.59} & 0.87 & 1.31 & 0.91 & 1.06 & 0.94 & 0.76 & 0.83 & 0.93 & 0.81 & 0.89 & 0.73 & 0.81 & 0.76 & 0.83 & 0.44 & 0.89 & 0.84 & 0.59 & 0.99 & 0.45 & 0.67 & \textbf{0.85} \\
GPT-oss             & 0.63 & 0.83 & 0.95 & 1.03 & 0.96 & 1.00 & 0.96 & 0.71 & 0.94 & 0.95 & 0.79 & 0.90 & 0.72 & 0.89 & 0.94 & 0.85 & 0.60 & 0.85 & 0.94 & 1.43 & 0.83 & 0.56 & 0.71 & 0.88 \\
Sutra               & 0.55 & 0.74 & 0.93 & 2.09 & 0.92 & 0.89 & 0.96 & 0.68 & 0.92 & 0.91 & 0.67 & 0.94 & 0.65 & 0.92 & 0.84 & 0.82 & 0.24 & 0.51 & 0.91 & \textbf{0.26} & 1.10 & 0.47 & 0.59 & 0.90 \\
Sarvam              & 0.99 & 0.66 & 0.91 & 1.50 & 1.00 & 1.27 & 1.13 & 0.64 & 0.85 & 1.19 & 0.62 & 0.99 & 0.65 & 2.19 & 0.72 & 0.96 & 0.24 & 0.54 & 0.93 & 1.45 & 3.63 & 0.45 & 0.56 & 4.25 \\
\hdashline
MUTANT-Indic & \textbf{0.45} & 0.60 & \textbf{0.65} & \textbf{0.94} & \textbf{0.78} & \textbf{0.85} & \textbf{0.82} & \textbf{0.54} & \textbf{0.68} & \textbf{0.80} & \textbf{0.53} & \textbf{0.76} & \textbf{0.50} & \textbf{0.91} & \textbf{0.61} & \textbf{0.67} & \textbf{0.18} & \textbf{0.45} & \textbf{0.66} & 0.45 & \textbf{0.72} & \textbf{0.38} & \textbf{0.44} & 0.86 \\
\bottomrule
\end{tabular}
}
\caption{NSL score ($\downarrow$)  comparison for Indic focused and Good Indic support tokenizers across  languages here. MUTANT-Indic performs best in 23 of 24 languages. An extended version in Table \ref{tab:nsl-scores-transposed} (Appendix).}
\label{tab:nsl-scores-main}
\end{table*}

\subsection{Evaluation Framework}
\label{subsec:eval-framework}
We develop a modular evaluation framework supporting HuggingFace\footnote{\url{https://github.com/huggingface/tokenizers}}, SentencePiece\footnote{\url{https://github.com/google/sentencepiece}}, and TikToken\footnote{\url{https://github.com/openai/tiktoken}} tokenizers along with a comprehensive set of intrinsic metrics (Appendix \ref{app:metrics}), including Fertility score, Normalized Sequence Length (NSL), Rényi entropy and efficiency, and Bytes Per Token (BPT). All metrics are computed at the line level and aggregated to the language level. 
To assess tokenizer behavior in Indic use cases, we construct an evaluation set spanning 22 Indic languages, English, and code.
Table~\ref{tab:eval-stats} reports dataset statistics: text volume, number of lines, and average words per line per language. We will release both the evaluation dataset and the framework for reproducible benchmarking and fair comparison of multilingual tokenizers.


\section{Experiments and Results}
\label{sec:experiments}

In our work, we compare MUTANT-Indic, trained using our recipe, against 9 tokenizers, comprising: \textit{i) Indic-focused tokenizers:} tokenizers designed primarily for Indian languages: Sutra \citep{tamang2024evaluating} and Sarvam-2B \citep{sarvam2024sarvam1} (referred as Sarvam). \textit{ii) Good Indic support tokenizers:} multilingual tokenizers with demonstrated capabilities for Indic languages: Gemma-3-27B-it \citep{team2025gemma} (referred as Gemma-3), GPT-OSS \citep{openai2025gptoss120bgptoss20bmodel} and LLaMA-4 \citep{meta2025llama4}. \textit{iii) General tokenizers:} tokenizers of widely-used general-purpose LLMs: Qwen3-32B \citep{yang2024qwen3} (referred as Qwen-3), LLaMA-3.2-1B \citep{dubey2024llama3}, Mistral-Nemo \citep{jiang2024mistralnemo} and DeepSeek-R1 \citep{deepseek2025deepseeekr1}.

\subsection{Intrinsic Evaluation of Tokenizers}
\label{sec:results}

We evaluate tokenization quality using four complementary intrinsic metrics that capture efficiency, granularity, and information utilization: (i) \textbf{Fertility score} \citep{rust2021good, scao2022bloom}, measuring subword fragmentation; (ii) \textbf{Normalized Sequence Length (NSL)} \citep{dagan2024getting}, quantifying relative compression against a base tokenizer; (iii) \textbf{Rényi entropy and efficiency} \citep{zouhar2023tokenization}, assessing information density and vocabulary utilization; and (iv) \textbf{Bytes per token} \citep{kocetkov2022stack}, reflecting memory and storage efficiency. All metrics are reported as micro-averages per line at the language level. Formal definitions are provided in Appendix~\ref{app:metrics}.

As shown in Table~\ref{tab:word-tok-ratio}, MUTANT-Indic achieves state-of-the-art fertility scores across most 
languages, consistently yielding the lowest token-to-word ratios among nine tokenizers with strong Indic support (see Appendix Table~\ref{tab:big_fertility_scores} for extended comparisons). This indicates substantially reduced fragmentation, particularly for morphologically rich scripts. We also evaluate our Stage-1 tokenizer (trained with BPE on the same data and LLaMA-4 pretokenization). As shown in Table~\ref{tab:stage-comparisons} in Appendix, our Stage-1 tokenizer, already achieves state-of-the-art (SOTA) performance across most Indic languages. Bytes-per-token results (Table~\ref{tab:bytes-per-token-transposed}) show that MUTANT-Indic achieves higher values across languages, reflecting more information-dense tokens and improved sequence compactness. Normalized Sequence Length scores (Table~\ref{tab:nsl-scores-main}) further confirms that MUTANT-Indic produces shorter tokenized sequences relative to baseline tokenizers, indicating superior compression efficiency. Finally, Rényi entropy and efficiency results (Table~\ref{tab:reyni-metrics}) demonstrate that MUTANT-Indic achieves consistently higher efficiency across languages, reflecting effective and balanced utilization of the vocabulary. Collectively, these intrinsic results establish MUTANT-Indic as a robust and efficient tokenizer across diverse scripts and languages.




\begin{table}[htbp]
\centering
\small
\resizebox{0.75\columnwidth}{!}{%
\begin{tabular}{lrr}
\toprule
Dataset & LLaMA-4 & MUTANT-Indic \\
\midrule
\textbf{English Benchmarks} \\
HellaSwag      & 0.353 & \textbf{0.357} \\
CommonsenseQA  & \textbf{0.206} & 0.204 \\
OpenBookQA     & 0.216 & \textbf{0.218} \\
Winogrande     & 0.504 & \textbf{0.510} \\
GSM8K          & 0.016 & \textbf{0.018} \\
ARC Easy       & 0.623 & \textbf{0.630} \\
ARC Challenge  & 0.291 & \textbf{0.292} \\
MMLU           & \textbf{0.252} & 0.249 \\
DROP           & \textbf{0.048} & 0.036 \\
\hdashline
Average        & \textbf{0.279} & \textbf{0.279} \\
\midrule
\textbf{Indic Benchmarks} \\
Indic COPA              & 0.544 & \textbf{0.556} \\
Indic Sentiment         & 0.524 & \textbf{0.551} \\
Indic XNLI              & \textbf{0.347} & 0.346 \\
Indic Paraphrase        & 0.534 & \textbf{0.539} \\
MILU (Indic Multi-turn LU) & \textbf{0.261} & 0.258 \\
ARC Challenge (Indic)   & 0.236 & \textbf{0.244} \\
TriviaQA (Indic)        & \textbf{0.268} & 0.262 \\
\hdashline
Average                 & 0.388 & \textbf{0.394} \\
\bottomrule
\end{tabular}}
\caption{Extrinsic evaluation of MUTANT-Indic vs LLaMA-4 on \textit{English} and \textit{Indic benchmarks}.}

\label{tab:benchmarks}
\end{table}

\subsection{Extrinsic Evaluation}

\label{subsec:model-training} 
We also evaluate the downstream model performance (see Table \ref{tab:benchmarks}) by pretraining LLaMA-3.2 1B models using two tokenizers: i) MUTANT-Indic, our proposed tokenizer optimized for morphologically meaningful segmentation in Indic and multilingual settings,
and (ii) LLaMA-4 tokenizer, chosen for comparable vocabulary size and widespread use. Both models were trained on the same dataset in iso-compute setting to ensure a fair comparison. More details in the Appendix \ref{app:imp}. We find that our tokenizer shows competitive performance across the English and Indic benchmarks. We additionally report Bits-per-Character in Appendix~\ref{app:bpb_eval}, which remains meaningful even near random accuracy. We also trained a model using the Stage-1 tokenizer, which also attains strong downstream performance. Table \ref{tab:benchmarks-s1} in Appendix shows that the Stage-1 tokenizer itself constitutes a strong and competitive baseline.

The pretraining corpus (Table~\ref{tab:data_distribution} in Appendix) balances coverage and domain diversity. It combines web-scale sources (Nemotron CC) for general context with structured data including MegaMath \citep{zhou2025megamath}, StackV2 \citep{lozhkov2024starcoder}, synthetic generations, and books. Indic-language content constitutes roughly $20\%$ of the corpus, drawn from Indic CC, Wikipedia, and Sangraha Verified \citep{khan2024indicllmsuite}, providing sufficient signal to evaluate cross-lingual and morphologically rich representation quality.



\begin{table}[htbp]
\centering
\resizebox{0.9\linewidth}{!}{
    \begin{tabular}{lcccccc}
    \toprule
    & Gemma-3 & GPT-OSS & LLaMA-4 & Sarvam & Sutra & MUTANT-Indic \\
    \midrule
    Entropy $\downarrow$ & 20.70 & 20.81 & 21.09 & 20.71 & 20.62 & \textbf{20.42} \\
    Efficiency $\uparrow$ & 0.22 & 0.19 & 0.14 & 0.21 & 0.23 & \textbf{0.28} \\
    \bottomrule
    \end{tabular}
    }
    \caption{Rényi's Entropy and Efficiency across top Indic tokenizers. Higher efficiency indicates better balance between vocabulary capacity and token usage.}
    \label{tab:reyni-metrics}
\end{table}

\begin{table}[htbp]
\centering
\resizebox{0.68\linewidth}{!}{
    \begin{tabular}{lrr}
    \toprule
    Model & TTFT (ms) $\downarrow$ & OTPT (tokens/s) $\uparrow$ \\
    \midrule
    LLaMA-4 & 19.17 $\pm$ 0.15 & 117.99 \\
    MUTANT-Indic     & \textbf{18.98 $\pm$ 0.36} & \textbf{169.42} \\
    \bottomrule
    \end{tabular}
}
\caption{Inference latency comparison of 1B models trained with LLaMA-4 and MUTANT-Indic tokenizers.}
\label{tab:model-latency}
\end{table}

\begin{table}[htbp]
\centering
\resizebox{\linewidth}{!}{
\begin{tabular}{c*{10}{c}}
\hline
Metric & ar & bn & deva & en & gu & ka & ml & pa & ta & te \\
\hline
Data size (MB)  & 106 & 396 & 2200 & 3590 & 124 & 644 & 580 & 307 & 616 & 617 \\
Percentage      & 1.12 & 4.18 & 23.25 & 37.94 & 1.31 & 6.81 & 6.13 & 3.24 & 6.51 & 6.52 \\
Vocab \% & 2.69 & 6.32 & 20.89 & 32.92 & 2.38 & 7.82 & 6.76 & 4.68 & 7.04 & 8.50 \\
\hline
\end{tabular}
}
\caption{Script-specific training data size (Total corpus size 9.4 GB) and resulting vocabulary \% distribution.}
\label{tab:unified-hypothesis}
\end{table}

\subsection{Impact on latency and throughput}

\label{sec:inference-latency}


Next, we evaluate how tokenization impacts end-to-end model efficiency. We train two 1B-parameter models under identical conditions:
one with our tokenizer and one with the LLaMA tokenizer of similar vocabulary size. We then evaluate inference efficiency over 200 samples spanning Indic languages and English, with varying input lengths. Latency\footnote{\url{https://tinyurl.com/4e7nh7c8}}
 was measured using standard metrics, including Time-To-First-Token (TTFT), Output Throughput (OTPT), and Input Sequence Length (ISL), across 200 instances ( See Appendix \ref{app:lat-otpt} for details) with 5 warm-up requests and results averaged over 10 runs. Experiments were done on 8 H100 GPUs using Triton Inference Server as backend, with a maximum generation limit of 256 new tokens. Our tokenizer yields clear efficiency gains (Table \ref{tab:model-latency}) which stem from
 improved compression: shorter token sequences encode more information per token, thereby lowering per-request computation without compromising expressivity. Overall, this demonstrates that tokenizer design directly shapes not only pretraining efficiency but also real-world deployment latency, making it a critical factor for practical model performance.

\subsection{Vocabulary Allocation: Explicit vs. Corpus-Driven}
\label{sec:vocab_alloc}
We compare explicit vocabulary merging with corpus-driven joint training under script-aware budget constraints. Explicitly merging script-specific tokenizers leads to fragmented segmentation and higher fertility due to cross-script interference (Table~\ref{tab:individual-vs-merged-selected}). In contrast, corpus-driven joint training naturally aligns vocabulary allocation with data distribution (Table \ref{tab:unified-hypothesis}), achieving lower fertility
and outperforming merged and public baselines (Table \ref{tab:word-tok-ratio}). These results indicate that joint, corpus-driven training is a more effective and scalable strategy.



        
    
    



\begin{table}[htbp]
\centering
\resizebox{0.9\linewidth}{!}{%
\begin{tabular}{l*{7}{c}}
\toprule
Tokenizer & as & bn & hi & mai & mr & san & te \\
\midrule
Individual & 2.05 & 2.13 & \textbf{1.21} & \textbf{1.35} & 1.75 & \textbf{2.49} & \textbf{1.40} \\
Merged & 2.32 & 2.14 & 1.55 & 1.57 & 1.73 & 2.79 & 1.95 \\
\hdashline
MUTANT-Indic & \textbf{1.85} & \textbf{1.74} & 1.23 & 1.58 & \textbf{1.63} & 2.59 & 1.88 \\
\bottomrule
\end{tabular}}
\caption{Fertility comparison ($\downarrow$) between individual script tokenizers and the merged tokenizer across selected Indic languages. Lower values are better.}
\label{tab:individual-vs-merged-selected}
\end{table}

\begin{table}[htbp]
\centering
\small
\resizebox{0.99\columnwidth}{!}{%
\begin{tabular}{lcc}
\toprule
Dataset & w/ Original LLaMA-4 & w/ MUTANT-Indic \\
\midrule
\textbf{English Benchmarks} \\
Winogrande     & 0.60 & \textbf{0.61} \\
GSM8K          & \textbf{0.05} & \textbf{0.05} \\
ARC Challenge  & \textbf{0.40} & 0.39 \\
MMLU           & \textbf{0.32} & 0.29 \\
\hdashline
Average        & \textbf{0.34} & \textbf{0.34} \\
\midrule
\textbf{Indic Benchmarks} \\
Indic COPA        & \textbf{0.58} & 0.56 \\
Indic Sentiment   & 0.82 & \textbf{0.85} \\
Indic XNLI        & \textbf{0.35} & 0.34 \\
Indic Paraphrase  & \textbf{0.57} & 0.53 \\
\hdashline
Average            & \textbf{0.58} & 0.57 \\
\bottomrule
\end{tabular}}
\caption{Performance comparison on English and Indic benchmarks in Continual Pretraining (CPT) setting of LLaMA-3.2 1B model (best scores in bold).}
\label{tab:retok}
\end{table}




\begin{table*}[htbp]
\centering
\small
\setlength{\tabcolsep}{3pt}
\renewcommand{\arraystretch}{1.1}
\resizebox{\textwidth}{!}{%
\begin{tabular}{l*{25}{r}}
\toprule
Tokenizer & as & bn & brx & code & doi & eng & gom & gu & hi & kas & kn & mai & ml & mni & mr & nep & or & pa & san & sat & snd & ta & te & urd & Avg \\
\midrule
Single-stage (200K) & 1.86 & 1.76 & 2.05 & 1.75 & 1.62 & 1.37 & 2.20 & 1.86 & 1.39 & 1.39 & 2.19 & 1.61 & \textbf{2.29} & 2.30 & 1.66 & 1.67 & 1.69 & 1.49 & 2.68 & \textbf{3.61} & 1.56 & 2.12 & 1.88 & 1.54 & 1.90 \\
Two-stage (180K/200K) & \textbf{1.85} & \textbf{1.74} & \textbf{2.04} & \textbf{1.47} & \textbf{1.45} & \textbf{1.12} & \textbf{2.17} & \textbf{1.77} & \textbf{1.23} & \textbf{1.21} & 2.19 & \textbf{1.58} & 2.30 & \textbf{2.28} & \textbf{1.63} & \textbf{1.62} & \textbf{1.65} & \textbf{1.39} & \textbf{2.59} & 3.72 & \textbf{1.45} & 2.12 & 1.88 & \textbf{1.44} & \textbf{1.83} \\
\bottomrule
\end{tabular}%
}
\caption{Fertility score ($\downarrow$) for single vs two-stage curriculum training (same evaluation data as Section~\ref{subsec:eval-framework}).}


\label{tab:two-stage-vs-one-stage}
\end{table*}

\begin{table*}[htbp]
\centering
\small
\resizebox{\textwidth}{!}{%
\begin{tabular}{c*{25}{c}}
\toprule
Parameter & as & bn & brx & code & doi & eng & gom & gu & hi & kas & kn & mai & ml & mni & mr & nep & or & pa & san & sat & snd & ta & te & urd & Average\\
\midrule
\textbf{Data size} \\
1G  & 3.02 & 2.32 & 2.71 & 1.62 & 1.64 & 1.33 & 1.97 & 1.62 & 1.50 & 1.43 & 2.16 & 1.85 & 2.83 & 2.62 & 1.72 & 2.13 & 1.68 & 1.50 & 2.46 & 13.02 & 1.43 & 1.92 & 1.82 & 1.91 & 2.42 \\
5G  & 1.71 & 1.93 & 2.58 & 1.63 & 1.58 & 1.33 & 2.18 & 1.72 & 1.40 & 1.36 & 2.04 & 1.57 & 2.43 & 2.28 & 1.68 & 1.48 & 1.61 & 1.57 & 2.48 & 4.74 & 1.30 & 2.02 & 1.87 & 1.43 & 1.91 \\
10G & 1.83 & 1.74 & 1.99 & 1.54 & 1.56 & 1.33 & 2.17 & 1.83 & 1.36 & 1.36 & 2.15 & 1.56 & 2.24 & 2.27 & 1.61 & 1.59 & 1.65 & 1.47 & 2.51 & 3.60 & 1.45 & 2.08 & 1.83 & 1.47 & \textbf{1.80} \\
25G & 1.75 & 1.84 & 2.56 & 1.62 & 1.57 & 1.33 & 2.15 & 1.78 & 1.39 & 1.36 & 2.04 & 1.56 & 2.32 & 2.23 & 1.67 & 1.47 & 1.63 & 1.55 & 2.45 & 3.92 & 1.31 & 2.01 & 1.86 & 1.34 & 1.86 \\
30G & 1.76 & 1.84 & 2.32 & 1.62 & 1.57 & 1.33 & 2.13 & 1.78 & 1.39 & 1.36 & 2.03 & 1.57 & 2.31 & 2.24 & 1.67 & 1.47 & 1.63 & 1.54 & 2.45 & 4.02 & 1.31 & 2.00 & 1.87 & 1.35 & 1.86 \\
50G & 1.72 & 1.82 & 2.25 & 1.60 & 1.57 & 1.34 & 2.14 & 1.82 & 1.39 & 1.36 & 2.03 & 1.58 & 2.28 & 2.22 & 1.69 & 1.49 & 1.64 & 1.52 & 2.44 & 4.54 & 1.31 & 2.01 & 1.87 & 1.34 & 1.87 \\
\midrule
\textbf{Transition point} \\
60 & 1.91 & 1.80 & 2.05 & 1.39 & 1.38 & 1.04 & 2.16 & 1.77 & 1.16 & 1.15 & 2.17 & 1.53 & 2.30 & 2.29 & 1.56 & 1.58 & 1.68 & 1.39 & 2.48 & 3.89 & 1.43 & 2.11 & 1.86 & 1.45 & 1.81 \\
75 & 1.91 & 1.79 & 2.05 & 1.41 & 1.38 & 1.04 & 2.16 & 1.77 & 1.16 & 1.15 & 2.16 & 1.53 & 2.30 & 2.28 & 1.56 & 1.58 & 1.68 & 1.39 & 2.47 & 3.91 & 1.43 & 2.10 & 1.86 & 1.45 & 1.81 \\
80 & 1.89 & 1.78 & 2.03 & 1.41 & 1.38 & 1.05 & 2.15 & 1.77 & 1.16 & 1.16 & 2.14 & 1.53 & 2.28 & 2.26 & 1.56 & 1.57 & 1.67 & 1.39 & 2.46 & 3.83 & 1.42 & 2.08 & 1.83 & 1.44 & 1.80 \\
85 & 1.87 & 1.77 & 2.01 & 1.43 & 1.39 & 1.06 & 2.13 & 1.76 & 1.17 & 1.16 & 2.13 & 1.53 & 2.26 & 2.25 & 1.56 & 1.56 & 1.66 & 1.39 & 2.46 & 3.78 & 1.42 & 2.07 & 1.82 & 1.44 & 1.80 \\
90  & 1.85 & 1.74 & 2.04 & 1.47 & 1.45 & 1.12 & 2.17 & 1.77 & 1.23 & 1.21 & 2.19 & 1.58 & 2.30 & 2.28 & 1.63 & 1.62 & 1.65 & 1.39 & 2.59 & 3.72 & 1.45 & 2.12 & 1.88 & 1.44 & 1.83 \\
95 & 1.85 & 1.75 & 1.98 & 1.47 & 1.42 & 1.10 & 2.13 & 1.74 & 1.21 & 1.20 & 2.12 & 1.53 & 2.23 & 2.24 & 1.56 & 1.56 & 1.66 & 1.41 & 2.46 & 3.68 & 1.43 & 2.06 & 1.81 & 1.44 & 1.79 \\
\midrule
\textbf{Vocab size} \\
162K/180K & 1.89 & 1.78 & 2.08 & 1.48 & 1.47 & 1.13 & 2.21 & 1.80 & 1.24 & 1.22 & 2.22 & 1.60 & 2.35 & 2.27 & 1.65 & 1.65 & 1.68 & 1.42 & 2.62 & 3.84 & 1.48 & 2.16 & 1.91 & 1.47  & 1.86 \\
180K/200K & 1.85 & 1.74 & 2.04 & 1.47 & 1.45 & 1.12 & 2.17 & 1.77 & 1.23 & 1.21 & 2.19 & 1.58 & 2.30 & 2.27 & 1.63 & 1.62 & 1.65 & 1.39 & 2.59 & 3.72 & 1.45 & 2.12 & 1.88 & 1.44 & 1.82 \\
202K/225K & 1.81 & 1.70 & 1.99 & 1.44 & 1.43 & 1.10 & 2.14 & 1.72 & 1.20 & 1.19 & 2.14 & 1.55 & 2.24 & 2.21 & 1.59 & 1.59 & 1.60 & 1.36 & 2.55 & 3.59 & 1.41 & 2.08 & 1.82 & 1.41 & 1.79 \\
225K/250K & 1.78 & 1.67 & 1.95 & 1.42 & 1.42 & 1.09 & 2.11 & 1.69 & 1.19 & 1.17 & 2.10 & 1.53 & 2.20 & 2.17 & 1.57 & 1.57 & 1.57 & 1.34 & 2.52 & 3.45 & 1.38 & 2.04 & 1.77 & 1.38  & 1.75\\
\bottomrule
\end{tabular}%
}
\caption{Ablation of tokenizer training data size (in GB), transition point (as a $\%$ of vocab size $200$K), vocab size ($t=90\%$) and its impact on fertility score ($\downarrow$). We use the same evaluation data as described in Section~\ref{subsec:eval-framework}.}
\label{tab:data_size_ablation}
\end{table*}

\subsection{Quality Analysis: Undertrained Tokens}
\label{sec:glitch}
We analyze under-trained ``Glitch'' tokens in our tied-embedding LLaMA-3.2-1B models trained with both the MUTANT-Indic tokenizer and a comparable BPE tokenizer of similar vocabulary size trained on the same corpus. Both tokenizers share the first 90\% of the vocabulary. The MUTANT-Indic tokenizer switches to multi-word training for the last 10\% whereas the base BPE tokeniser continues standard subword training. Following \citet{land2024fishingmagikarpautomaticallydetecting} to construct a reference for unused embeddings, we introduced a small set of dummy tokens into the vocabulary that have zero occurrences in the training data. Their embeddings were averaged to obtain a mean reference vector. We then retrieve the top-$K$ nearest neighbors (cosine distance), which represent potential ``glitch'' tokens~\citep{geiping2024coercing}. As shown in Figure \ref{fig:glitch-token-trend} (in the Appendix \ref{app:glitch-tokens}), the MUTANT-Indic tokenizer produces far fewer such glitch tokens than the base BPE tokenizer. These results suggest that incorporating multi-words promotes more efficient utilization of the vocabulary, while purely subword-based tokenizers overfit in the long tail, yielding a higher proportion of under-trained tokens. 

\subsection{Can we replace Opensource model tokenizer with MUTANT-Indic?}


Following ReTok~\citep{gu2024retokreplacingtokenizerenhance}, we replace the tokenizer of a pre-trained LLaMA-3.2-1B model (denoted \textsc{LLaMA-3.2-Orig})~\citep{grattafiori2024llama} with MUTANT-Indic (referred as LLaMA-3.2-MUTANT-Indic). 
Let $V_{\text{orig}}$ and $V_{\text{MUTANT-Indic}}$ denote their corresponding vocabularies. For a token $t \in V_{\text{MUTANT-Indic}}$, we initialize its embedding $E_{\text{init}}(t)$ as: if $t \in V_{\text{orig}} \cap V_{\text{MUTANT-Indic}}$, then $E_{\text{init}}(t) = E_{\text{orig}}(t)$, its embedding from the pretrained model, otherwise, if $t \in V_{\text{MUTANT-Indic}} \setminus V_{\text{orig}}$ and decomposes under the original tokenizer into $(t_1,\ldots,t_k)$, then $E_{\text{init}}(t) = \tfrac{1}{k}\sum_{i=1}^k E_{\text{orig}}(t_i)$.  

We then continually pretrained the LLaMA-3.2-MUTANT-Indic model, keeping just the embedding and LM head layers trainable, on a 40B-token corpus comprising English, Indic, code, and mathematics (see Appendix for details). 
As seen in Table~\ref{tab:retok}, the LLaMA-3.2-MUTANT-Indic model performs competitively with the original LLaMA-3.2-ORIG. This suggests that, in addition to pretraining-from-scratch settings, an optimized multilingual tokenizer, such as MUTANT-Indic, could also be leveraged in opensource models through CPT (Continual Pretraining~\citep{chen2024towards}) leading to  
significant throughput gains (see Table~\ref{tab:model-latency}). 

\section{Ablation studies}
\label{sec:ablation}

\paragraph{Two-Stage vs. One-Stage: Controlling Vocabulary}
\label{subsec:two-stage-vs-one-stage}


Recently, BoundlessBPE \citep{schmidt2024boundless} also explored a one-stage training paradigm in which pre-tokenization is governed by a fixed regular expression, enabling the direct learning of multiword units in a single pass for English language. Our approach instead introduces a two-stage procedure. We replicate the one-stage setup of BoundlessBPE using its released regex and compare against our strategy. 
Our method consistently achieves lower fertility across the top 10 Indic languages and English (Table~\ref{tab:two-stage-vs-one-stage}).
Overall, the comparison highlights a clear trade-off: while one-stage methods capture surface-level patterns indiscriminately, our two-stage design balances efficiency and linguistic integrity by decoupling subword and multiword learning.


\paragraph{Dataset Size}

Similar to \citet{reddy2025much}, we study the effects of scaling training data in Stage 1 of our training. Table~\ref{tab:data_size_ablation} shows that the performance plateaus after 10G of data.

\paragraph{Transition Point}

We ablate the transition point \(t\) (Section  \ref{sec:stage-1-algo}) at which training shifts from subword to cross-word merges. Varying \(t\) reveals a clear trade-off: early transitions favor frequent multiword expressions but weaken morphological coverage, while late transitions preserve subwords at the cost of longer sequences. Across Indic and non-Indic languages, intermediate values of 85-90\% \(t\) yields the best balance, improving token efficiency and cross-lingual consistency (Table~\ref{tab:data_size_ablation}).

\paragraph{Vocabulary Size}

Vocabulary size strongly influences tokenization-model efficiency with trade-offs. Smaller vocabularies yield finer subword units that generalize well to unseen words but with increased compute costs. Larger vocabularies shorten sequences by encoding frequent forms as single tokens, but waste capacity on rare items, inflate embeddings and softmax layers~\citep{shazeer2017outrageouslylargeneuralnetworks}, and have bias toward high-resource languages, hurting multilingual balance. With the same transition point at 90\%, we found no significant impact on fertility scores beyond 200K 
(Table \ref{tab:data_size_ablation}).

\paragraph{Effect of Normalization}

Prior works show that unicode normalization is crucial for multilingual settings~\citep{karthika2025multilingualtokenizationlensindian}, 
which reduces
token fragmentation and inflated vocabulary size. Table \ref{table:norm} (in Appendix) shows that NFKC yielded marginal but consistent gains by unifying character forms. Accordingly, we adopt NFKC to reduce variability and improve tokenizer robustness.

\section{Conclusion}
\label{sec:conclusion}

In this work, we revisit tokenization as a central design choice for multilingual LLMs and introduce MUTANT, a principled and general recipe for multilingual tokenizer training. We demonstrate the effectiveness of this framework with MUTANT-Indic, which achieves SOTA performance across Indic languages through systematic analyses of vocabulary size, data distribution, language-specific pre-tokenization, and multi-stage learning. Extensive evaluation across intrinsic efficiency metrics, downstream tasks, ablations, and inference latency, in both pretraining-from-scratch and continual pretraining settings, highlights consistent gains in efficiency and deployment cost. We will additionally publicly release our evaluation framework to support reproducibility and future research.


\newpage
\section*{Limitations}

While MUTANT-Indic achieves strong intrinsic efficiency gains and practical throughput improvements, our study has several limitations. First, although we evaluate across 22 Indic languages, English, and code, the analysis is restricted to these language families, and extending the framework to other low-resource or non-Indic languages remains future work. Second, downstream evaluations are limited to models up to 1B parameters due to compute constraints; while prior scaling-law results suggest these gains extrapolate to larger models, we do not validate this empirically. Third, although we analyze morphology-aware pre-tokenization, we do not integrate it into the final tokenizer due to its high inference latency, leaving efficient implementations to future work. Finally, while our evaluation framework standardizes intrinsic metrics, it does not capture all linguistic dimensions relevant to downstream tasks.

\section*{Ethical Considerations}
\paragraph{Ethics Statement}
This work focuses on the responsible development of multilingual tokenization methods for Indian languages. We did not collect or utilize any sensitive or Personally Identifiable Information (PII). AI tools (like ChatGPT) were used to refine text writing and section phrasing in the paper. All external datasets, libraries, and tools employed in this work are appropriately acknowledged through citations. Since the study did not involve personal, medical, or otherwise sensitive information, formal IRB approval was not required. Throughout the process, we aimed to minimize biases that could disadvantage low-resource languages. We provide our exhaustive study to advance the development of inclusive and efficient multilingual language models.

\paragraph{Reproducibility Statement}
To promote transparency and reproducibility, we will release the artifacts publicly to benchmark performance of Indian tokenizers, along with detailed documentation. Detailed records of experimental setups, hyperparameters, and evaluation protocols are maintained to allow replication of our results with the implementation details in the Appendix. In addition, we provide ablation studies to facilitate fair benchmarking and enable future research on Indian and multilingual tokenization.

\section*{Acknowledgement}
We sincerely thank the Krutrim leadership for their continued support in conducting this research.
We would also like to thank Aditya Kallappa, Gautam Rajeev, Guduru Manoj, Shaharukh Khan and Neel Rachamalla
for their help in different parts of this work, and the Research team at Krutrim for their valuable
suggestions and insightful discussions.


\bibliography{custom}


\null
\vfill
\eject

\newpage
\appendix
\section*{Appendix}

\section{Language Details}
\label{app:iso}

We provide more details about the 22 languages and corresponding scripts considered in our work in Table \ref{tab:language_coverage}. Additionally, ISO code mapping is also provided in Table \ref{tab:lang-mapping}. 

\begin{table*}[htbp]
\centering

\resizebox{\textwidth}{!}{%
\begin{tabular}{lll}
\toprule
Family & Script & Languages \\
\midrule

Indo-Aryan & Devanagari & Hindi, Marathi, Maithili, Dogri, Konkani, Sanskrit, Nepali, Kashmiri \\
           & Bengali (bn) & Assamese, Bengali  \\
           & Gurmukhi (pa) & Punjabi  \\
           & Arabic (ar) & Urdu, Sindhi \\

Dravidian  & Kannada (kn) & Kannada \\
           & Malayalam (ml) & Malayalam  \\
           & Tamil (ta) & Tamil \\
           & Telugu (te) & Telugu \\

Tibeto-Burman & Devanagari & Bodo \\
              & Meitei Mayek & Manipuri (Meitei Mayek script) \\

Austroasiatic & Ol Chiki (sat) & Santali \\

\bottomrule
\end{tabular}
}
\caption{Linguistic composition of the 22 scheduled Indian languages analyzed in this work, with their corresponding scripts.}
\label{tab:language_coverage}
\end{table*}

\begin{table}[htbp]
\centering
\renewcommand{\arraystretch}{1.2}
\setlength{\tabcolsep}{6pt}
\resizebox{\linewidth}{!}{%
\begin{tabular}{l l | l l | l l}
\toprule
Code & Language & Code & Language & Code & Language \\
\midrule
as & Assamese   & bn  & Bengali    & brx & Bodo       \\
doi & Dogri     & gu  & Gujarati   & hi  & Hindi      \\
kn  & Kannada   & ks  & Kashmiri   & gom & Konkani    \\
mai & Maithili  & ml  & Malayalam  & mni & Manipuri   \\
mr  & Marathi   & ne  & Nepali     & or  & Odia       \\
pa  & Punjabi   & san & Sanskrit   & sat & Santali    \\
snd & Sindhi    & ta  & Tamil      & te  & Telugu     \\
ur  & Urdu      &     &            &     &            \\
\bottomrule
\end{tabular}}
\caption{Mapping of ISO codes to corresponding 22 Indic languages.}

\label{tab:lang-mapping}
\end{table}

\section{Implementation}
\label{app:imp}

\subsection{Tokenizer implementation}
\label{appendix:tokenizer-imp}
Our training code for the tokenizer is based on the open implementation of SuperBPE\footnote{\url{https://github.com/PythonNut/superbpe/tree/main}} using HuggingFace library~\citep{jain2022hugging}. 
We first curate a multilingual corpus of $\sim$50G from OLMo~\citep{olmo20252olmo2furious}, Wikipedia\footnote{\url{https://en.wikipedia.org/wiki/}}, books, PDFs, Common Crawl, and the Sangraha dataset \citep{khan2024indicllmsuite} where we sample our tokenizer training data from varied sources to get enough representation of the different languages individually. English data was mostly sampled from OLMo and code data from Stackv2~\citep{lozhkov2024starcoder}. 
The common crawl corpus was language-identified using fastText LID,
deduplicated using MinHash, filtered for boilerplate and HTML artifacts, normalized using Unicode NFKC and script-validated to reduce cross-script contamination, similar to FineWeb \citep{penedo2024fineweb}.
The final MUTANT-Indic tokenizer uses a shared vocabulary of $200$K tokens, distributed across language scripts and is trained on $10$GB of multilingual high quality data.

In our work, we also explored merging tokenizers based on the default priority based BPE in SentencePiece\footnote{\url{https://github.com/google/sentencepiece}}. While we explored implementing the multi-word two stage curriculum in the SentencePiece, we found that it was not trivial. On the other hand, HuggingFace showed issues with the merging strategy. We thus relied on different implementations for different approaches. 

\subsection{Training details}
\label{appendix:train-details}
We provide more details about our training setup as discussed in Section \ref{subsec:model-training}. 
Each model was trained for 50B tokens under matched hyperparameters (learning rate, batch size, training steps), aligning FLOPs to isolate tokenizer effects. The evaluation was performed using \texttt{lm-eval-harness} \citep{eval-harness} across standard English benchmarks (MMLU, GSM8K, Winogrande, TriviaQA, HellaSwag, ARC, OpenBookQA, CommonsenseQA, DROP) and Indic benchmarks (IndicCOPA, IndicSentiment, IndicXParaphrase, IndicXNLI \citep{doddapaneni2023leavingindiclanguagebehind}, ARC Challenge Indic \citep{sarvamai_arcc_indic}, and MILU \cite{verma2024milu}). We report EM for GSM8K and TriviaQA, F1 for DROP, and Accuracy for other benchmarks. Shot settings were fixed per task: $25$-shot for ARC/ARC Challenge Indic, $10$-shot for HellaSwag, $5$-shot for MMLU, GSM8K, and TriviaQA, and zero-shot for the remainder. This setup allows a direct assessment of how tokenizer design influences pretraining efficiency, semantic representation, and generalization across English and Indic tasks.

\begin{table*}[htbp]
\centering
\scriptsize
\resizebox{\linewidth}{!}{%
\begin{tabular}{lccccccccc}
\hline
Tokenizer & Architecture & Parameters & Data Size (B ) & Learning Rate & Train Steps & Context Length & Batch Size & Vocab Size \\
\hline
LLaMA-4  & LLaMA-3.2 & 1B & 53.24 & $5 \times 10^{-5}$ & 68000 & 4096 & 192 & 201134  \\
MUTANT-Indic          & LLaMA-3.2 & 1B & 53.18 & $5 \times 10^{-5}$ & 68000 & 4096 & 192 & 200008  \\

\hline
\end{tabular}}
\caption{Pretraining configuration for different tokenizers. }

\label{tab:pretrain_config}
\end{table*}

\begin{table}[h!]
\centering
\scriptsize
\setlength{\tabcolsep}{6pt}
\renewcommand{\arraystretch}{1.2}
\begin{tabular}{lrrr}
\hline
\textbf{Category} & \textbf{Sources} & \textbf{Percentage (\%)} & \textbf{Token Count (B)} \\
\hline
Web & Nemotron CC & 30 & 15 \\
Math & MegaMath & 15 & 7.5 \\
Code & StackV2 & 15 & 7.5 \\
Synthetic & New Generations & 10 & 5 \\
Books & Archive & 10 & 5 \\
Indic & Indic CC & 8 & 4 \\
Indic & Indic Wiki & 4 & 2 \\
Indic & Sangraha Verified & 8 & 4 \\
\hline
\textbf{Total} & & 100 & 50 \\
\hline
\end{tabular}
\caption{Pretraining corpus distribution across domains and token count. Indic content is emphasized to reflect multilingual objectives.}

\label{tab:data_distribution}
\end{table}

\section{Additional Discussion}

\subsection{Mismatch Between Loss and Task Performance}
Although toknizers, incorporating multi-word often show slightly higher loss~\citep{liu2025superbpe} during training compared to models using traditional atomic tokenizers like SentencePiece/BPE, this does not necessarily translate to worse downstream performance. We hypothesize that this is due to two complementary factors. First, the introduction of longer or multi-word tokens such as ``to the" or ``as well as" increases the number of semantically overlapping candidates, making the model’s prediction space less sharply peaked. This means the model may distribute probability across several plausible completions (e.g., ``to", ``to the", ``to be"), thereby lowering the maximum assigned probability to the correct token and inflating the cross-entropy loss. In contrast, other BPE tokenizers often yield only one atomic candidate for such function words, allowing sharper predictions with lower loss. Second, MUTANT-Indic tokenizes text into fewer, more meaningful units, so when computing the average loss per token, each mistake contributes more heavily to the total. As a result, although the model learns more compact and generalizable representations, its token-level loss appears higher. This creates a divergence between model loss and real-world task accuracy, indicating that traditional loss curves may underrepresent the representational efficiency and practical utility of compositional tokenizers like MUTANT-Indic.

\subsection{Morphologically grounded token splitting}
\label{sec:morph}
We investigate the impact of incorporating morphological information into tokenization for Indic languages~\citep{brahma2025morphtok}. The approach involves pre-processing text with a morphology analyzer to segment words into morphemes prior to training. This experiment focuses on languages in the Devanagari script.

\begin{figure*}[htbp]
    \centering
    \begin{minipage}{0.48\linewidth}
        \centering
        \includegraphics[width=\linewidth]{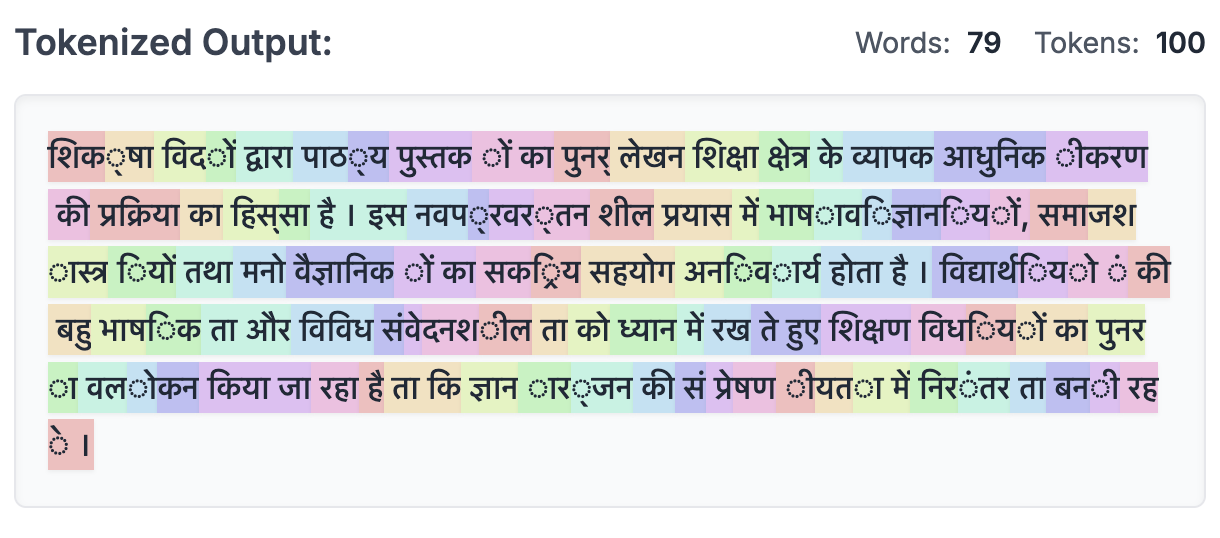}
        \caption{Tokenized output of morph-aware tokenizer}
        \label{fig:reyni-efficiency-eng}
    \end{minipage}
    \hfill
    \begin{minipage}{0.48\linewidth}
        \centering
        \includegraphics[width=\linewidth]{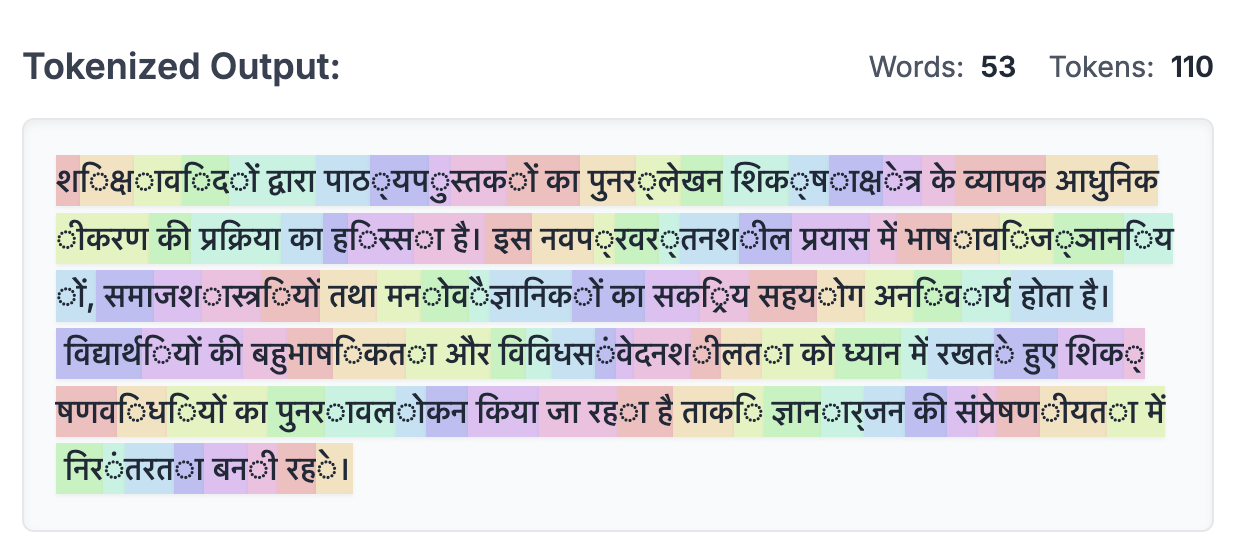}
        \caption{Tokenized output of non morph-aware tokenizer}
        \label{fig:reyni-efficiency-indic}
    \end{minipage}
\end{figure*}

We compare two variants: Tokenizer A, trained on raw text, and Tokenizer B, trained on morphologically segmented text using morphology analyzer ~\citep{kunchukuttan2020indicnlp}. At inference time, Tokenizer B requires the same pre-processing for consistency. Tokenizer B exhibits more semantically coherent multi-words, reflecting meaningful morpheme combinations (Figure~\ref{fig:reyni-efficiency-eng}, \ref{fig:reyni-efficiency-indic}). This promotes better generalization across related forms and reduces the raw token-to-word ratio, as morpheme-based units are more compressible. Sample outputs (Figures~\ref{fig:reyni-efficiency-eng}, \ref{fig:reyni-efficiency-indic}) illustrate the contrast between surface-level splits and linguistically aligned segmentations.

Despite these gains, we do not adopt this approach in our final tokenizer. The primary limitation is latency, as the pipeline requires both language identification and morphological analysis. For completeness, we evaluated a Hindi morphology-aware tokenizer augmented with a morphological analyzer~\citep{kunchukuttan2020indicnlp} combined with language identification (LID)\footnote{\url{https://pypi.org/project/langdetect/}}. We performed inference on approximately 4-5 MB of Hindi text and measured throughput over 10 runs (with 5 warm-up runs), comparing against our MUTANT-Indic tokenizer. Our tokenizer achieved 194K tokens/sec, whereas the morphology-aware tokenizer achieved 90K tokens/sec, representing a 53.28\% reduction in throughput. This slowdown arises entirely from the additional LID and morphology-analysis stages, underscoring the efficiency advantages of our approach even when compared to linguistically informed baselines.
Extending robust analyzers across all Indic languages also introduces engineering overhead and brittle dependencies. Nevertheless, morphology-aware tokenization remains a promising direction if fast, reliable analyzers become widely available.

\begin{figure}[htbp]
    \centering
    \includegraphics[width=0.65\linewidth]{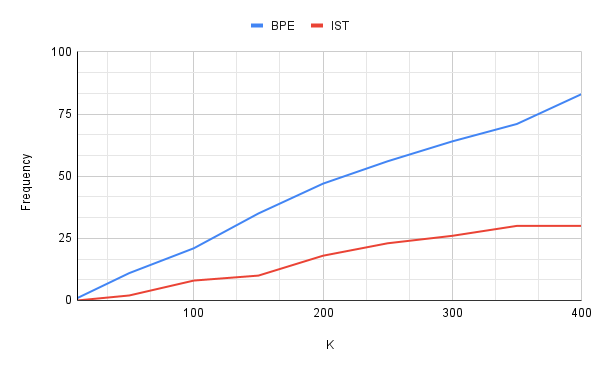}
    \caption{Trend of potential glitch tokens in upper 20K of vocabulary for different K.}
    \label{fig:glitch-token-trend}
\end{figure}

\subsection{More on Glitch tokens}
\label{app:glitch-tokens}
For each tokenizer, we vary $K \in \{10, 50, 100, 150, \dots, 400\}$ to select the top-$K$ embeddings closest to a reference vector derived from artificially unused tokens in the vocabulary~\citep{land2024fishingmagikarpautomaticallydetecting,geiping2024coercing}. For the MUTANT-Indic tokenizer, we count the number of multi-word tokens within the top-$K$. For the BPE variant, we count tokens with IDs $>180{,}000$, which corresponds to the upper 20K of the vocabulary. Both tokenizers share the first 180K IDs; the difference lies in how the final 20K IDs are utilized: MUTANT-Indic allocates this space for frequent multi-word tokens, while the BPE tokenizer continues learning subwords. This design choice allows MUTANT-Indic to more effectively utilize the tail of the vocabulary for meaningful units, whereas the BPE tokenizer exhibits overfitting in low-frequency subwords.
The trend of these counts across different top-$K$ values is visualized in Figure~\ref{fig:glitch-token-trend}. As $K$ increases, the fraction of multi-word tokens in MUTANT-Indic remains low but stable, while the BPE variant consistently shows a higher fraction of under-trained subwords, indicating overfitting in the residual vocabulary space.

\subsection{More on Latency and Throughput Evaluation}
\label{app:lat-otpt}

To obtain reliable latency and throughput measurements, we constructed a 200-example multilingual inference set intended to approximate realistic LLM workloads. The set contains diverse sentence-completion style prompts representative of common generation patterns. We include 20 inputs per language across English and nine major Indic languages, ensuring balanced coverage of script diversity, lexical variation, and syntactic complexity. Table \ref{tab:eval200} presents the token-length distribution of these examples under both the LLaMA-4 tokenizer and our MUTANT-Indic, allowing a controlled comparison of inference efficiency across tokenization schemes.



\begin{table}[htbp]
\centering
\caption{Token-length statistics for the 200-example inference set.
We report min, p75, p90, p99, maximum, and average token lengths.}
\label{tab:eval200}
\resizebox{\linewidth}{!}{%
\begin{tabular}{lrrrrrr}
\toprule
\textbf{Tokenizer} & \textbf{min} & \textbf{p75} & \textbf{p90} & \textbf{p99} & \textbf{max} & \textbf{avg} \\
\midrule
Llama-4 & 288 & 805 & 1157 & 2583 & 2869 & 784 \\
MUTANT-Indic & 178 & 440 & 541 & 654 & 676 & 379 \\
\bottomrule
\end{tabular}}
\end{table}

\subsection{Details About Baseline Tokenizers}
Tokenizer fertility is shaped by multiple factors including training data distribution, vocabulary construction, and underlying algorithmic choices, yet publicly available documentation on these aspects is often limited. Table~\ref{tab:baseline-tokenizers} summarizes the vocabulary sizes, training methodologies, and any disclosed data distributions for all baseline tokenizers considered in our study.

\begin{table}[htbp]
\centering
\resizebox{\linewidth}{!}{
\begin{tabular}{llll}
\toprule
\textbf{Tokenizer} & \textbf{Vocab Size} & \textbf{Training Algorithm / Framework} & \textbf{Data Distribution} \\
\midrule
DeepSeek-R1 & 128K & BPE (undisclosed variant) & Not publicly disclosed \\
Gemma-3 & 262K & SentencePiece & 140+ languages \\
GPT-OSS & 200K & o200k\_harmony (TikToken variant) & Not publicly disclosed \\
LLaMA-3.2-1B & 128K & BPE / SentencePiece-based & Not publicly disclosed \\
LLaMA-4 & 200K & BPE & Not fully disclosed \\
Mistral-Nemo & 131K & Tekken tokenizer (TikToken-based) & 100+ languages; multilingual + code \\
Qwen-3 & 151K & Byte-level BPE & Not publicly disclosed \\

Sarvam & 68K & Not publicly disclosed & Not publicly disclosed \\
Sutra & 256K & SentencePiece (unigram/BPE hybrid) & Balanced multilingual; uniform sampling \\
\bottomrule
\end{tabular}
}
\caption{Summary of baseline tokenizers and publicly available training details.}
\label{tab:baseline-tokenizers}

\end{table}


\subsection{Bits-per-Character Evaluation on Multiple Choice Benchmarks}
\label{app:bpb_eval}

In addition to task-level accuracy, we evaluate the language modeling quality of the models using \textit{bits-per-character} (BPC) measured on the gold answer continuations of benchmark examples. Following \citet{heineman2025signal}, we compute BPC as:

\begin{equation}
\text{BPC}(y|x) = - \frac{1}{|y| \log 2} \sum_{t=1}^{|y|} \log P(y_t \mid x, y_{<t}),
\end{equation}

where $x$ denotes the prompt, $y$ is the gold continuation (correct answer text), and $|y|$ is the number of character in the continuation. Lower values indicate better modeling of the target continuation.

This metric provides a continuous measure of modeling quality that is independent of the discrete multiple-choice selection. It allows comparison of tokenizers through the likelihood assigned to the correct answers in benchmark datasets, where accuracy is near chance.

\paragraph{English Benchmarks}

Table~\ref{tab:bpb_english} reports BPC values on several English multiple-choice benchmarks. Across these tasks, the model trained with the MUTANT-Indic tokenizer consistently achieves lower BPC compared to the model trained with the LLaMA-4 tokenizer, indicating improved likelihood modeling of the correct answer continuations.

\begin{table}[h]
\centering
\resizebox{\linewidth}{!}{
\small
\begin{tabular}{lccccc}
\toprule
\textbf{Tokenizer} & \textbf{ARC-C} & \textbf{ARC-E} & \textbf{OBQA} & \textbf{HellaSwag} & \textbf{CSQA} \\
\midrule
MUTANT-Indic & \textbf{1.0284} & \textbf{1.2368} & \textbf{1.5774} & \textbf{0.9867} & \textbf{2.3445} \\
LLaMA-4      & 1.2575 & 1.6037 & 1.8345 & 0.9983 & 2.7398 \\
\bottomrule
\end{tabular}}
\caption{Bits-per-character (lower is better) on English multiple-choice benchmarks. ARC-C: ARC-Challenge, ARC-E: ARC-Easy, OBQA: OpenBookQA, CSQA: CommonsenseQA.}
\label{tab:bpb_english}
\end{table}

\paragraph{Indic Benchmark}

We further evaluate BPC on translated ARC-Challenge benchmarks across ten Indic languages. Results are shown in Table~\ref{tab:bpb_indic}. The MUTANT-Indic tokenizer generally produces lower BPC values across the evaluated languages, suggesting improved modeling efficiency for the gold answer continuations.

\begin{table}[t]
\centering
\resizebox{\linewidth}{!}{
\small
\begin{tabular}{lcccccccccc}
\toprule
\textbf{Tokenizer} & \textbf{bn} & \textbf{gu} & \textbf{hi} & \textbf{kn} & \textbf{ml} & \textbf{mr} & \textbf{or} & \textbf{pa} & \textbf{ta} & \textbf{te} \\
\midrule
MUTANT-Indic & 1.2747 & 1.3480 & 1.3066 & 1.2113 & 1.0298 & 1.3205 & 1.1975 & 1.2054 & 0.9317 & 1.1535 \\
LLaMA-4      & 1.2756 & 1.4275 & 1.3254 & 1.3193 & 1.2264 & 1.4368 & 1.3694 & 1.3935 & 1.0508 & 1.2889 \\
\bottomrule
\end{tabular}}
\caption{Bits-per-character (lower is better) on ARC-Challenge across Indic languages.}
\label{tab:bpb_indic}
\end{table}

\subsection{Stage-2 Regex Boundary Constraints}
\label{sec:regex_exp}

Stage~1 pre-tokenization relies on established regex-based segmentation strategies commonly used in multilingual tokenizers (e.g., GPT-2 and LLaMA-style rules). These patterns separate punctuation, normalize whitespace boundaries, and produce stable token units prior to subword learning.

Stage~2 instead applies additional regex constraints that regulate subword merges during tokenizer training. The objective is to enable learning of frequent multi-word expressions while preventing unstable token formations.

Specifically, the Stage~2 regex enforces: (i) sentence boundary preservation, preventing merges across sentence delimiters; (ii) explicit handling of Indic punctuation such as the Devanagari markers `|'' and `||'' as hard boundaries; and (iii) controlled segmentation of numeric sequences.

\section{Metrics Definitions}
\label{app:metrics}

Here, we discuss the different intrinsic metrics used in our evaluation framework. 

\subsection{Token-to-Word Ratio}
\label{app:metrics-twr}

The Token-to-word ratio measures the average number of tokens required to represent a single word. It captures the degree of segmentation induced by a tokenizer and is particularly informative for morphologically rich languages where excessive fragmentation increases sequence length. We report this metric to evaluate whether tokenizers balance compact representations with sufficient linguistic coverage.

\subsection{Bytes-per-token}
\label{app:metrics-bpt}

Bytes-per-token quantifies the average number of raw text bytes contained in a token. Since scripts differ substantially in character set size and encoding, this metric provides a language-agnostic measure of efficiency. Higher values indicate that tokens encode more information per unit, which reduces sequence length. We include this metric to enable direct comparison of tokenizers across writing systems.

\subsection{Normalized Sequence Length}
\label{app:metrics-nsl}
Normalized sequence length measures the average length of tokenized sequences relative to a chosen base tokenizer. Instead of reporting absolute sequence lengths, this metric highlights how much longer or shorter sequences become when compared to an established reference. It enables fairer cross-tokenizer comparisons since raw lengths can vary significantly across languages and corpora. A normalized value greater than one indicates that the tokenizer produces longer sequences than the baseline, while a value less than one reflects more compact tokenization. We include this metric to directly assess relative efficiency in sequence compression.

\subsection{Reyni's Efficiency}
\label{app:metrics-re}
Rényi’s entropy measures the uncertainty of token distributions induced by a tokenizer, extending Shannon entropy by allowing different orders to emphasize frequent or rare tokens. A tokenizer with a very large vocabulary may contain many infrequent tokens that are poorly utilized, while a very small vocabulary forces overuse of common tokens. Entropy therefore reflects how effectively the vocabulary is allocated. To complement this, Rényi’s efficiency normalizes entropy with respect to vocabulary size, providing a scale-invariant view of how well the vocabulary capacity is utilized. Together, these metrics characterize both the distributional balance of tokens and the comparative efficiency of different vocabulary scales.

\section{Extended Results}
\label{appendix:more-results}

In the main paper, due to space constraints, we limited the number of tokenizers presented. Here, we provide an extended list including all of our baseline tokenizers. Table \ref{tab:nsl-scores-transposed} and \ref{tab:big_fertility_scores} provides the NSL scores and fertility scores of the other tokenizers considered in our work. Table \ref{table:norm} shows the fertility scores for different normalization techniques in our MUTANT-Indic tokenizer. Additionally, we also compare the performance of two models trained with only Stage-1 and the 2-stage curriculum tokenizer in Table \ref{tab:benchmarks-s1}. 

\begin{table*}[htbp]
\centering
\small
\caption{Comparison of NSL scores (Base LLaMA-4) for different tokenizers across all languages.}
\resizebox{\textwidth}{!}{
\begin{tabular}{lrrrrrrrrrrrrrrrrrrrrrrrr}
\toprule
Tokenizer ($\downarrow$) & as & bn & brx & code & doi & eng & gom & gu & hi & kas & kn & mai & ml & mni & mr & nep & or & pa & san & sat & snd & ta & te & urd \\
\midrule
DeepSeek-R1         & 0.83 & 0.97 & 1.25 & 1.03 & 1.29 & 0.98 & 1.28 & 1.48 & 1.59 & 1.29 & 1.41 & 1.34 & 1.52 & 0.99 & 1.49 & 1.61 & 0.67 & 1.41 & 1.19 & 0.69 & 1.34 & 0.82 & 1.34 & 1.21 \\
Gemma-3             & 0.63 & 0.59 & 0.87 & 1.31 & 0.91 & 1.06 & 0.94 & 0.76 & 0.83 & 0.93 & 0.81 & 0.89 & 0.73 & 0.81 & 0.76 & 0.83 & 0.44 & 0.89 & 0.84 & 0.59 & 0.99 & 0.45 & 0.67 & 0.85 \\
GPT-OSS             & 0.63 & 0.83 & 0.95 & 1.03 & 0.96 & 1.00 & 0.96 & 0.71 & 0.94 & 0.95 & 0.79 & 0.90 & 0.72 & 0.89 & 0.94 & 0.85 & 0.60 & 0.85 & 0.94 & 1.43 & 0.83 & 0.56 & 0.71 & 0.88 \\
LLaMA-3.2-1B        & 1.90 & 2.71 & 1.08 & 1.02 & 1.36 & 0.99 & 1.22 & 2.91 & 1.47 & 1.36 & 3.30 & 1.16 & 3.25 & 1.92 & 1.41 & 1.44 & 1.48 & 2.45 & 1.19 & 1.34 & 1.33 & 2.11 & 3.01 & 1.58 \\
LLaMA-4             & 1.00 & 1.00 & 1.00 & 1.00 & 1.00 & 1.00 & 1.00 & 1.00 & 1.00 & 1.00 & 1.00 & 1.00 & 1.00 & 1.00 & 1.00 & 1.00 & 1.00 & 1.00 & 1.00 & 1.00 & 1.00 & 1.00 & 1.00 & 1.00 \\
Mistral-Nemo        & 1.00 & 0.95 & 1.06 & 1.15 & 1.07 & 1.06 & 1.09 & 1.09 & 1.12 & 1.08 & 0.91 & 1.08 & 0.95 & 0.95 & 1.13 & 1.21 & 1.57 & 0.98 & 1.04 & 1.34 & 1.20 & 0.63 & 0.64 & 0.95 \\
Qwen-3              & 1.68 & 2.37 & 1.78 & 1.11 & 1.85 & 1.03 & 1.72 & 2.59 & 2.65 & 2.16 & 2.69 & 1.97 & 2.57 & 1.72 & 2.35 & 2.47 & 1.19 & 2.37 & 1.92 & 0.96 & 1.37 & 1.63 & 2.45 & 1.63 \\
Sutra               & 0.55 & 0.74 & 0.93 & 2.09 & 0.92 & 0.89 & 0.96 & 0.68 & 0.92 & 0.91 & 0.67 & 0.94 & 0.65 & 0.92 & 0.84 & 0.82 & 0.24 & 0.51 & 0.91 & 0.26 & 1.10 & 0.47 & 0.59 & 0.90 \\
Sarvam              & 0.99 & 0.66 & 0.91 & 1.50 & 1.00 & 1.27 & 1.13 & 0.64 & 0.85 & 1.19 & 0.62 & 0.99 & 0.65 & 2.19 & 0.72 & 0.96 & 0.24 & 0.54 & 0.93 & 1.45 & 3.63 & 0.45 & 0.56 & 4.25 \\
\hdashline
MUTANT-Indic     & 0.45 & 0.60 & 0.65 & 0.94 & 0.78 & 0.85 & 0.82 & 0.54 & 0.68 & 0.80 & 0.53 & 0.76 & 0.50 & 0.91 & 0.61 & 0.67 & 0.18 & 0.45 & 0.66 & 0.45 & 0.72 & 0.38 & 0.44 & 0.86 \\
\bottomrule
\end{tabular}
}
\label{tab:nsl-scores-transposed}
\end{table*}

\begin{table*}[htbp]
\centering
\caption{Fertility scores across tokenizers and languages. Lower is better.}

\renewcommand{\arraystretch}{1.3} 
\resizebox{\textwidth}{!}{
\begin{tabular}{lcccccccccccccccccccccccccccc}
\toprule
Tokenizer ($\downarrow$) & as & bn & brx & code & doi & eng & gom & gu & hi & kas & kn & mai & ml & mni & mr & nep & or & pa & san & sat & snd & ta & te & urd \\
\midrule
DeepSeek-R1 & 3.54 & 2.88 & 4.23 & 1.53 & 2.66 & 1.34 & 3.68 & 4.92 & 3.02 & 2.49 & 6.01 & 3.21 & 7.95 & 2.67 & 4.17 & 3.97 & 7.13 & 4.48 & 5.07 & 6.12 & 2.82 & 4.92 & 6.13 & 2.17 \\
Gemma3 & 2.65 & 1.69 & 2.84 & 1.79 & 1.69 & 1.39 & 2.60 & 2.50 & 1.47 & 1.48 & 3.34 & 1.91 & 3.45 & 2.07 & 2.03 & 2.03 & 4.42 & 2.83 & 3.37 & 5.16 & 2.03 & 2.50 & 2.94 & 1.44 \\
GPT-OSS & 2.66 & 2.41 & 3.17 & 1.51 & 1.89 & 1.33 & 2.73 & 2.37 & 1.72 & 1.58 & 3.34 & 2.01 & 3.51 & 2.41 & 2.61 & 2.10 & 6.26 & 2.71 & 3.89 & 13.01 & 1.76 &3.18 & 3.13 & 1.51 \\
Llama-3.2-1B & 8.44 & 8.08 & 3.64 & 1.51 & 2.92 & 1.35 & 3.46 & 9.95 & 2.74 & 2.70 & 14.44 & 2.79 & 16.26 & 5.31 & 3.90 & 3.52 & 15.68 & 7.88 & 4.86 & 12.15 & 2.85 & 12.25 & 13.68 & 2.73 \\
LLaMA-4 & 4.40 & 2.93 & 3.34 & 1.46 & 2.00 & 1.34 & 2.84 & 3.37 & 1.83 & 1.72 & 4.23 & 2.28 & 4.95 & 2.73 & 2.79 & 2.46 & 10.51 & 3.23 & 4.12 & 9.04 & 2.13 & 5.87 & 4.53 & 1.76 \\
Mistral-Nemo & 4.28 & 2.82 & 3.52 & 1.75 & 2.12 & 1.41 & 3.08 & 3.63 & 2.05 & 1.82 & 3.84 & 2.48 & 4.82 & 2.67 & 3.10 & 2.97 & 16.92 & 3.04 & 4.34 & 12.16 & 2.51 & 3.67 & 3.71 & 1.65 \\
Qwen3-32B & 7.47 & 7.11 & 6.10 & 1.68 & 4.05 & 1.41 & 5.08 & 8.87 & 4.86 & 3.70 & 11.48 & 4.53 & 12.77 & 4.76 & 6.56 & 6.10 & 12.37 & 7.60 & 8.04 & 8.81 & 2.95 & 9.69 & 11.10 & 2.90 \\
Sarvam-2B & 4.24 & 1.91 & 2.92 & 2.14 & 1.85 & 1.66 & 3.01 & 2.11 & 1.53 & 1.91 & 2.53 & 2.11 & 3.19 & 4.60 & 1.94 & 2.35 & 2.43 & 1.67 & 3.78 & 13.07 & 7.62 & 2.49 & 2.63 & 7.93 \\
Sutra & 2.12 & 2.07 & 3.06 & 2.12 & 1.78 & 1.17 & 2.68 & 2.15 & 1.62 & 1.48 & 2.71 & 2.08 & 3.10 & 2.40 & 2.18 & 2.01 & 2.24 & 1.50 & 3.76 & 2.03 & 2.23 & 2.58 & 2.77 & 1.55 \\
\hdashline
MUTANT-Indic & 1.85 & 1.74 & 2.04 & 1.47 & 1.45 & 1.12 & 2.17 & 1.77 & 1.23 & 1.21 & 2.19 & 1.58 & 2.30 & 2.28 & 1.63 & 1.62 & 1.65 & 1.39 & 2.59 & 3.72 & 1.45 & 2.12 & 1.88 & 1.44 \\

\bottomrule
\end{tabular}%
}
\label{tab:big_fertility_scores}
\end{table*}

\begin{table*}[htbp]
\centering
\resizebox{\textwidth}{!}{%
\begin{tabular}{lrrrrrrrrrrrrrrrrrrrrrrrr}
\toprule
 Regex   & as & bn & brx & code & doi & eng & gom & gu & hi & kas & kn & mai & ml & mni & mr & nep & or & pa & san & sat & snd & ta & te & urd \\
\midrule
MUTANT-Indic (Stage-1) & 1.83 & 1.74 & 1.99 & 1.54 & 1.56 & 1.33 & 2.17 & 1.83 & 1.36 & 1.36 & 2.15 & 1.56 & 2.24 & 2.27 & 1.61 & 1.59 & 1.65 & 1.47 & 2.51 & 3.60 & 1.45 & 2.07 & 1.83 & 1.47 \\
MUTANT-Indic (Stage-2) & 1.85 & 1.74 & 2.04 & 1.47 & 1.45 & 1.12 & 2.17 & 1.77 & 1.23 & 1.21 & 2.19 & 1.58 & 2.30 & 2.28 & 1.63 & 1.62 & 1.65 & 1.39 & 2.59 & 3.72 & 1.45 & 2.12 & 1.88 & 1.44 \\
\bottomrule
\end{tabular}%
}
\caption{Fertility scores ($\downarrow$) comparing our tokenizer after Stage 1 and Stage 2 curriculum. Stage-1 tokenizer corresponds to: cleaned corpus, standard BPE, LLaMA-4 regex, identical vocabulary and data size. The only difference between this baseline and our final MUTANT-Indic is the Stage-2 curriculum phase.}
\label{tab:stage-comparisons}
\end{table*}


\begin{table*}[htbp]
\centering
\scriptsize

\setlength{\tabcolsep}{3pt}
\renewcommand{\arraystretch}{1.1}
\resizebox{\textwidth}{!}{%
\begin{tabular}{lrrrrrrrrrrrrrrrrrrrrrrrr}
\toprule
Tokenizer & as & bn & brx & code & doi & eng & gom & gu & hi & kas & kn & mai & ml & mni & mr & nep & or & pa & san & sat & snd & ta & te & urd \\
\midrule
NFC    & 1.8520 & 1.7449 & 2.0412 & 1.4658 & 1.4520 & 1.1167 & 2.1741 & 1.7664 & 1.2250 & 1.2042 & 2.1845 & 1.5761 & 2.3025 & 2.2421 & 1.6273 & 1.6241 & 1.6464 & 1.3915 & 2.5859 & 3.7170 & 1.4515 & 2.1226 & 1.8754 & 1.4371 \\
NFD    & 1.8518 & 1.7454 & 2.0413 & 1.4665 & 1.4521 & 1.1168 & 2.1661 & 1.7667 & 1.2252 & 1.2044 & 2.1905 & 1.5765 & 2.3019 & 2.2487 & 1.6274 & 1.6246 & 1.6465 & 1.3917 & 2.5864 & 3.7170 & 1.4523 & 2.1227 & 1.8757 & 1.4377 \\
NFKC   & 1.8512 & 1.7430 & 2.0409 & 1.4647 & 1.4520 & 1.1155 & 2.1738 & 1.7644 & 1.2239 & 1.2041 & 2.1812 & 1.5762 & 2.2991 & 2.2327 & 1.6258 & 1.6234 & 1.6420 & 1.3884 & 2.5855 & 3.7172 & 1.4505 & 2.1200 & 1.8724 & 1.4369 \\
\bottomrule
\end{tabular}%
}
\caption{Fertility scores with NFC, NFD, NFKC  normalization for all languages.}
\label{table:norm}
\end{table*}

\begin{table*}[htbp]
\centering
\caption{ Comparison of downstream performance between MUTANT-Indic (Stage-1) and  MUTANT-Indic (Stage-2).}
\small
\setlength{\tabcolsep}{4pt}
\renewcommand{\arraystretch}{1.1}
\resizebox{\linewidth}{!}{%
\begin{tabular}{lrr|lrr}
\toprule
\multicolumn{3}{c|}{English Benchmarks\textbf{ }} & \multicolumn{3}{c}{Indic Benchmarks\textbf{ }} \\
\cmidrule(lr){1-3} \cmidrule(lr){4-6}
Dataset & MUTANT-Indic-Stage-1 & MUTANT-Indic-Stage-2 &
Dataset & MUTANT-Indic-Stage-1 & MUTANT-Indic-Stage-1 \\
\midrule
HellaSwag             & 0.348 & 0.357 & Indic COPA              & 0.556 & 0.556 \\
CommonsenseQA         & 0.193 & 0.204 & Indic Sentiment         & 0.557 & 0.551 \\
OpenBookQA            & 0.214 & 0.218 & Indic XNLI              & 0.366 & 0.346 \\
Winogrande            & 0.515 & 0.510 & Indic Paraphrase        & 0.562 & 0.539 \\
GSM8K                 & 0.021 & 0.018 & MILU (Indic Multi-turn LU) & 0.265 & 0.258 \\
ARC Easy              & 0.625 & 0.630 & ARC Challenge (Indic)   & 0.247 & 0.244 \\
ARC Challenge         & 0.279 & 0.292 & TriviaQA (Indic)        & 0.268 & 0.262 \\
MMLU                  & 0.255 & 0.249 & \multicolumn{2}{c}{} \\
DROP                  & 0.042 & 0.036 & \multicolumn{2}{c}{} \\
\midrule
Average      & 0.277 & 0.279 &
Average      & 0.403 & \textbf{0.394} \\
\bottomrule
\end{tabular}}
\label{tab:benchmarks-s1}
\end{table*}

\end{document}